\documentclass{article}

\usepackage{microtype}
\usepackage[dvipsnames]{xcolor}
\usepackage{graphicx}
\usepackage{subfigure}
\usepackage{booktabs} 
\newcommand{\citet}[1]{\citeauthor{#1} \shortcite{#1}}
\newcommand{\citep}{\cite}

\usepackage{amsmath,amssymb,amsfonts}
\usepackage{multirow}
\usepackage[shortlabels]{enumitem}
\usepackage[cal=cm]{mathalfa} 
\usepackage{subfigure}
\usepackage{subfloat}
\usepackage{microtype}      
\usepackage{amsmath}
\usepackage{multirow}
\usepackage{graphicx}
\usepackage{booktabs}
\usepackage{pgfplots}
\usepgfplotslibrary{fillbetween}
\pgfplotsset{compat=1.3}
\usepackage{hyperref}

\usepackage[accepted]{icml2020}

\icmltitlerunning{S-SGD: Symmetrical Stochastic Gradient Descent for Reaching Flat Minima in Deep Neural Network Training}

\begin{document}

\twocolumn[
\icmltitle{S-SGD: Symmetrical Stochastic Gradient Descent\\for Reaching Flat Minima in Deep Neural Network Training}



\icmlsetsymbol{equal}{*}

\begin{icmlauthorlist}
\icmlauthor{Wonyong Sung}{snu}
\icmlauthor{Iksoo Choi}{snu}
\icmlauthor{Jinhwan Park}{snu}
\icmlauthor{Seokhyun Choi}{snu}
\icmlauthor{Sungho Shin}{snu}

\end{icmlauthorlist}

\icmlaffiliation{snu}{Seoul National University, Seoul, Korea}
\icmlcorrespondingauthor{Wonyong Sung}{wysung@snu.ac.kr}


\vskip 0.3in
]
\printAffiliationsAndNotice{} 

\begin{abstract}
The stochastic gradient descent (SGD) method is the most widely used for deep neural network (DNN) training. However, this method requires the use of appropriate regularization techniques to prevent overfitting and improve the generalization capability.  Weight perturbations such as DropConnect and noise injection are widely used techniques. We propose an easy-to-use regularization method that adds symmetrical noises to the DNN weights. The proposed  Symmetrical-SGD (S-SGD) algorithm evaluates the loss surface at two separate points during training to avoid convergence to sharp minima. The training stability is considerably improved by injecting fixed-magnitude  symmetrical noise. The S-SGD method was applied to image classification using convolutional neural network models, end-to-end speech recognition with Gated ConvNet, and language modeling using recurrent neural network models. In all experiments, S-SGD outperformed conventional weight noise injection and dropout-based regularization techniques.   
\end{abstract}

\section{Introduction}\label{sec:intro}
Recently, there have been many studies on the shape of the loss function, and it has become clear that local minima that generalize well lie on wide valleys of the loss landscape, rather than in sharp, isolated minima \cite{hochreiter1997flat, keskar2016large}. In addition, the spectrum of Hessian matrix exhibits that well-generalized networks contain low eigenvalues of Hessian \cite{ghorbani2019investigation}.

The stochastic gradient descent (SGD) algorithm is widely used for deep neural network (DNN) training, but its convergent properties in non-convex optimizations are explained only in a probabilistic manner. Many types of research have been conducted to help SGD passing through saddle points and sharp minima by providing momentum or changing the learning rate in the weight updates~\cite{jastrzkebski2018finding,kleinberg2018alternative,chaudhari2016entropy,loshchilov2016sgdr,keskar2017improving,johnson2013accelerating,smith2017cyclical,jastrzkebski2018dnn,kingma2014adam}. In particular, SGD-based training demands the use of regularization techniques to prevent overfitting and improve the generalization capabilities. Popular regularization techniques include weight decay, Dropout, DropConnect, and noise injection to weights or gradients. \cite{zur2009noise,ho2008weight,murray1993synaptic,an1996effects,wen2018smoothout,jin2017escape,pmlr-v28-wan13, srivastava2014dropout}.

This study was strongly motivated by recent research on loss surface visualization and sharpness measurements \cite{li2018visualizing,keskar2016large}. Loss surface flatness can be visualized by inferring a DNN model after injecting random noise to the weights. A flat surface indicates that the loss does not appreciably change  when the weights are perturbed by the injected noise \cite{chaudhari2016entropy}. This suggests that we can improve the training of a DNN model while observing the loss surface or using the error term that depends on the surface flatness.  

In this study, we propose a symmetrical SGD (S-SGD) algorithm that computes the loss and flatness to reach flat minima in DNN training. The loss surface flatness is measured using two sets of weights formed by adding and subtracting the same noises to the current model in adaptation. When arriving at minima, we can derive that the gradient for weight updates also depends on the Hessian, or the flatness. To reduce the fluctuation of loss via perturbation, we add symmetrical noises of a fixed amount to the weights.  We compare this approach with the conventional SGD and weight noise injection-based methods using convolutional neural network (CNN) and recurrent neural network (RNN) models. 

This paper is organized as follows. In Section \ref{sec:relatedWorks}, we review related works on regularization and loss surface measurement methods. The proposed training algorithm is described in Section \ref{sec:algorithm}. Section \ref{sec:optimize} includes the hyperparameter optimization. The experimental results are shown in Section \ref{sec:exp}. Section \ref{sec:discussion} discusses this algorithm in comparison to other regularization techniques. Section \ref{sec:conclude} concludes the paper.

\section{Related Works}\label{sec:relatedWorks}

Many regularization techniques have been developed to ensure that DNNs do not overfit the training set. An early study showed that flat minima in the loss surface are closely related to good generalization \cite{hochreiter1997flat}. Recent studies also confirmed that reaching flat minima or wide valleys of loss landscape in DNN training provide robustness against weight and data disturbances, suggesting good generalization capabilities. Many training techniques have been developed to aid the SGD method to escape from sharp minima \cite{johnson2013accelerating,keskar2017improving,jastrzkebski2018finding}. One is a gradient update method, such as momentum \cite{qian1999momentum}, Adam \cite{kingma2014adam}, RMSProp \cite{Tieleman2012}, and Adagrad \cite{duchi2011adaptive}. Learning rate scheduling is known to be an effective technique for improving the generalization capability. These techniques include linear learning rate scaling \cite{smith2017don} and warm-up training \cite{loshchilov2016sgdr}.

Weight perturbation using noise modulation or injection has been studied for regularization for a long time. A well-known method is the Dropout that randomly drops some units during training \cite{srivastava2014dropout}. DropConnect is based on a similar idea, but it drops weights, not units, using random masks \cite{pmlr-v28-wan13}. Three types of noise injection methods have been studied; input data, weight, and gradient \cite{zur2009noise}. Although the concept dates back more than 20 years, research remains quite active \cite{murray1993synaptic,an1996effects}. Several recent studies can be found in \cite{wen2018smoothout, chaudhari2016entropy}. 

Measuring flatness of local minima can help to understand DNN training and optimization. Eigenvalues of Hessian characterize the loss surface sharpness. However, Hessian computation is not feasible because a DNN typically has millions of parameters. Instead, \cite{keskar2016large} proposed $\epsilon$-sharpness, which measures the maximum value within a distance $\epsilon$ from the local minimum. \cite{li2018visualizing} visualized loss surfaces by projecting `filter normalized' parameters to the space defined with random directions. This corresponds to adding weight noises of normalized magnitude many times and observing the loss surface. Recently, Entropy-SGD \cite{chaudhari2016entropy} was developed to construct a local-entropy-based objective function that favors well-generalized solutions lying in large flat regions of the loss landscape.

\section{Algorithm and Operation}\label{sec:algorithm}

In this section, we describe the conventional SGD, weight noise injection, and the proposed S-SGD algorithms.  Then, the convergence property of the S-SGD is analyzed. 

\subsection{Conventional SGD, Weight Noise Injection, and S-SGD Methods}

The conventional SGD method that operates with a batch size of $B$ updates the weights using Eq. \ref{eqn:SGD}, as follows:
\begin{align}
\mathbf{w}_{t+1} &= \mathbf{w}_{t} - \eta \sum_{i=1}^{B}\nabla L_{i}(\mathbf{w}_t), \label{eqn:SGD}
\end{align}

\noindent where $L_{i}$ is the loss computed with the $i$-th data in the batch and $\mathbf{w}_t$ are the trainable parameters. $\eta$ is proportionally scaled to the batch size $B$ \cite{zhu2018anisotropic, jastrzkebski2018finding}. When the batch size is very large, approximately $1/10$ of the total training data size, it is termed large batch training. The SGD method is based on the gradient descent, which was developed to solve convex optimization problems. However, the finite batch size introduces data-dependent variability or noise in inferring the loss function; as a result, the small-batch SGD method is known to be more effective in escaping from sharp minima than the large-batch SGD \cite{keskar2016large, goyal2017accurate}.   

The weight and gradient noise injection algorithms were developed to provide perturbation to the weights, which can be described as follows 
\begin{align}
\mathbf{w}_{t+1} &= \mathbf{w}_{t} - \eta \sum_{i=1}^{B}\nabla L_{i}(\widetilde{\mathbf{w}}_t), \label{eqn:perturb_SGD3}
\end{align}

\noindent where $\widetilde{\mathbf{w}}_t = \mathbf{w}_t + \mathbf{n}_{t}$, and $\mathbf{n}_{t}$ is the weight noise injected. Typically, the noise has uniform or Gaussian distributions \cite{wen2018smoothout}. The weight update is conducted using the derivative of the loss obtained with noise-injected weights. As the weight noise perturbs the network, the loss surface measured is blunt; thus, the training algorithm cannot properly recognize sharp minima, increasing the chance of skipping them. 

In this study, we propose a symmetric weight noise injection method, termed S-SGD method.  The proposed algorithm injects two symmetrical noises, $\mathbf{n}_{t}$ and $-\mathbf{n}_{t}$, to the weight, $\mathbf{w}_t$, to form $\widetilde{\mathbf{w}}_{t+}$ and $\widetilde{\mathbf{w}}_{t-}$. Forward and backward propagation operations are conducted using these two  weight sets and the gradients obtained are averaged and used for the weight updates. The magnitude or L2 norm of the injected noise, $\mathbf{n}_t$, is constant or slowly adapted and is proportional to the L2 norm of the weights for each layer. This scheme demands twice the amount of computation when compared to the conventional weight noise injection method. However, it shows very good ability in finding flat minima as presented in Section \ref{sec:optimize}.

\begin{figure}[t]
\centering
    \includegraphics[width=0.8\linewidth]{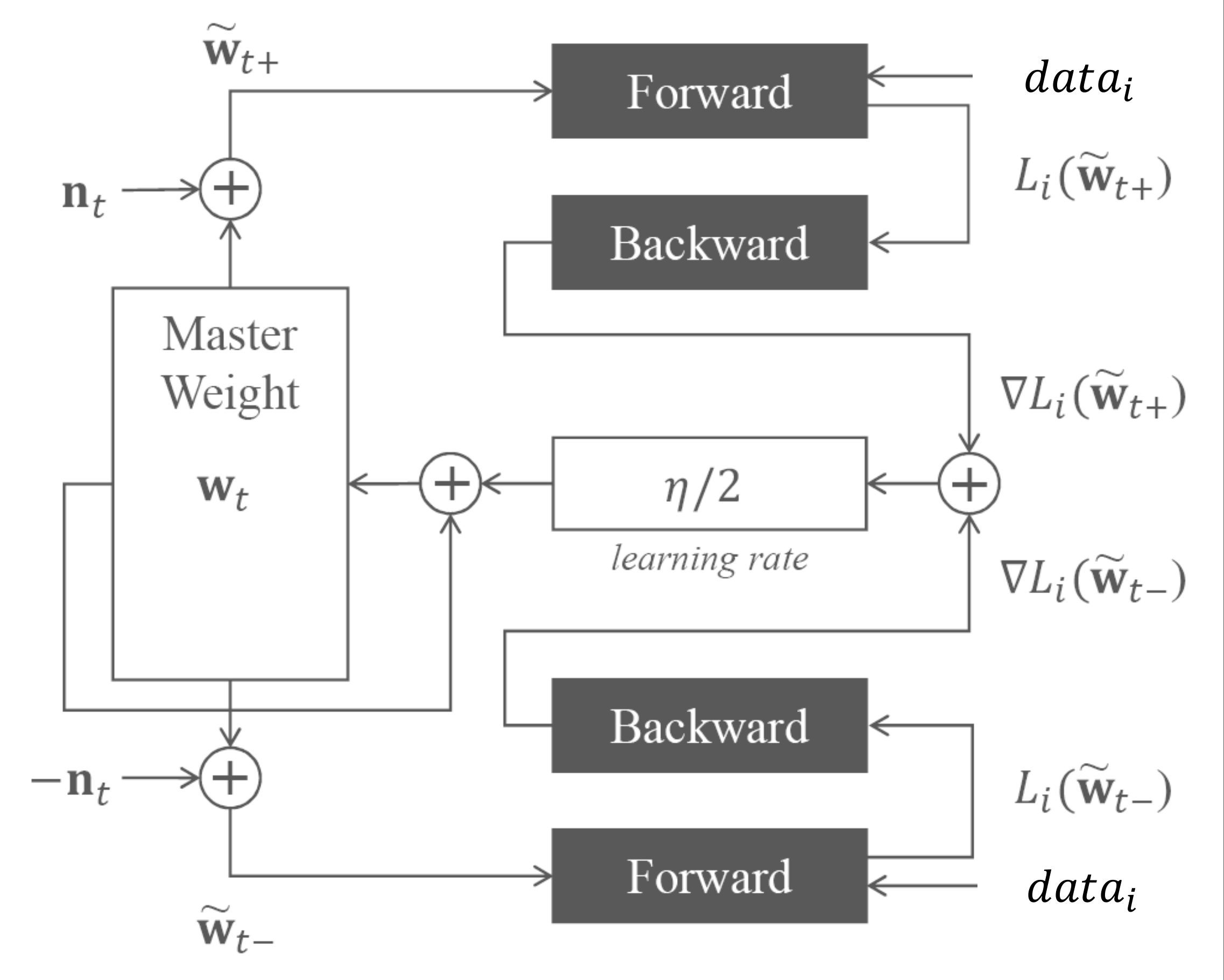}
\caption{The structure of Symmetrical SGD.}
\label{fig:fig_quant}  
\end{figure}

The update equation for S-SGD is as follows:
\begin{align}
\mathbf{w}_{t+1} &= \mathbf{w}_{t} - \eta \sum_{i=1}^{B}\frac{\nabla  (L_{i}(\widetilde{\mathbf{w}}_{t+}) + L_{i}(\widetilde{\mathbf{w}}_{t-}))}{2}, \label{eqn:perturb_SGD2}
\end{align}
The structure of the S-SGD is shown in \figurename~\ref{fig:fig_quant}. Two independent paths, one with $\widetilde{\mathbf{w}}_{t+}$  and the other with $\widetilde{\mathbf{w}}_{t-}$, are used for loss measurements and weight updates. Two symmetrical noises are added to the weights, and the magnitude of the noise is fixed or very slowly adjusted. The amount of injected noise is related to loss surface sharpness to escape. The weight update is conducted on the master weight only,  $\mathbf{w}_t$. 
In the next sub-section, we explain the S-SGD operation in more detail.

\subsection{S-SGD Operation}


\begin{figure}[t]
\begin{minipage}[b]{0.9\linewidth}
     \centering
    \pgfplotsset{compat=1.8, width=1.0\linewidth}
\begin{tikzpicture}
\begin{axis}[
    legend pos = north west,
    xmin=0,
    xmax=58,
    ymax=70,
    xmajorticks=false,
    xlabel={}, 
    ylabel={MCR (\%)}]
\addplot[red] table [x index=0, y index=2, col sep=tab]{data.txt};
\label{plot_one}
\addlegendentry{Train}
\addplot[blue] table [x index=0, y index=1, col sep=tab]{data.txt};
\addlegendentry{Test}
\draw [->, line width=0.5mm] 
    (axis cs: 8,0) -- (axis cs: 8,12);
\draw [->, line width=0.5mm] 
    (axis cs: 48,0) -- (axis cs: 48,8);
\draw [dash pattern= on 1pt off 1pt, line width=0.25mm, gray] 
    (axis cs: 8,0) -- (axis cs: 8,70);
\draw [dash pattern= on 1pt off 1pt, line width=0.25mm, gray] 
    (axis cs: 48,0) -- (axis cs: 48,70);
\draw[dash pattern= on 1pt off 1pt, gray, line width=0.25 mm] 
    (axis cs: 0,30.98) -- (axis cs:58,31.29);
\end{axis}
\draw node(S-SGD) at (0.9,-0.3){S-SGD};
\draw node(SGD) at (4.8,-0.3){SGD};
\end{tikzpicture}
    \end{minipage}
\caption{  1-D linear interpolation of solutions obtained by SGD and S-SGD. The misclassification ratio (MCR) of train and test sets are plotted.}
\label{fig:fig_ssgd}  
\end{figure}
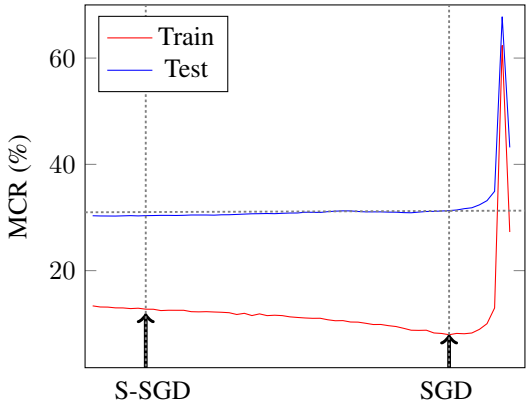


The S-SGD method updates weights using the gradients measured at two symmetrically perturbed weights, $\widetilde{\mathbf{w}}_{t+}$ and $\widetilde{\mathbf{w}}_{t-}$. The forward and backward procedure using the proposed method is identical to that of the conventional SGD method; a difference, however, is that both $\widetilde{\mathbf{w}}_{t+}$ and  $\widetilde{\mathbf{w}}_{t-}$ are used for forward and backward propagation.

For the simplicity of explanation, we assume to have only one variable $w$ instead of $\mathbf{w}$, which means that $w$ is a one-dimensional vector and the added noise $n$ is a positive constant.

The loss with the perturbed weight, $\widetilde{w}_{+} = w + n$, can be approximated as follows:
\begin{align}
  L(w+n) &= L(w) + (L(w+n)-L(w)) \\
      &= L(w) + \frac{(L(w+n)-L(w))}{n} \cdot n 
  \label{eqn:loss_approximate}
\end{align}

According to the mean-value theorem, we have the following equation.
\begin{align}
&L(w+n) = L(w) + \nabla L(w+ n_1 ) \cdot n,
\label{eqn:mean_value}
\end{align}
where $n_1$ is between 0 and $n$. 

At the same way,
\begin{align}
&L(w-n) = L(w) - \nabla L(w - n_2 ) \cdot n,
\label{eqn:mean_value2}
\end{align}
where $n_2$ is between 0 and $n$. 
\begin{align}
&\nabla L_{S-SGD} (w) = \frac{\nabla L(w+n) + \nabla L(w-n)}{2}\\
             &= \nabla L(w) + \frac{\nabla^2 L(w + n_1) - \nabla^2 L(w - n_2)}{2} \cdot n
\label{eqn:grad_approximate}
\end{align}

Note that the first term, $\nabla L(w)$, is the same as that of the derivative of SGD loss, while the second term is related to the second order derivatives.

In DNNs, $\mathbf{w}_{t}$ is a multi-dimensional variable and the noise $\textbf{n}_{t}$ is also a multi-dimensional random number. Note that $\textbf{n}_{t}$ has a constant magnitude, but is randomly generated for each weight update.

In Eq. \ref{eqn:mean_value}, we assume a derivative at $w+{n}_{1}$, which can easily be determined in the one-dimensional case.  However, in the multi-dimensional case, this is a derivative only at the direction of noise $\mathbf{n}$. As the dimension is as large as the number of parameters, it is not practical to evaluate the exact derivatives. Thus, we only measure the derivative in the direction of randomly generated noise or use random sampling. If the flatness of the loss function around the current weight in training is fairly smooth, we can assume to have a fairly good gradient estimate or the estimate of $\nabla L(\mathbf{w}_{t}+\mathbf{n}_{t})$ and $\nabla L(\mathbf{w}_{t}-\mathbf{n}_{t})$. Then, the $\nabla^2$ in Eq. \ref{eqn:grad_approximate} can be considered the approximation of the Hessian function obtained via random sampling. 

Eq. \ref{eqn:grad_approximate} suggests that if $\mathbf{w}$ is at the center of a flat minimum in which Hessian terms at $\mathbf{w}_{t}+\mathbf{n}_{t}$ and $\mathbf{w}_{t}-\mathbf{n}_{t}$ are similar, it becomes an optimum point to reach; in contrast, if the Hessian terms at $\mathbf{w}_{t}+\mathbf{n}_{1,t}$ and $\mathbf{w}_{t}-\mathbf{n}_{2,t}$ are considerably different or have the opposite signs, the weight updates continue even if $\nabla L(\mathbf{w}_{t})$ is zero.  

\figurename~\ref{fig:fig_ssgd} clearly shows the difference of solutions obtained by SGD and S-SGD. This is a one-dimensional misclassification ratio (MCR) function when the weights are perturbed along the direction of $\mathbf{w}_{SGD}$ and $\mathbf{w}_{S-SGD}$ This plot is obtained from the ResNet20 training using CIFAR-100 dataset. We can observe that the SGD solution corresponds to the point whose training loss or training MCR is the lowest. S-SGD finds the point whose loss is slightly higher than that of SGD, but its solution is moved toward the center of the flat minimum. The injected noise makes S-SGD maintain some distance from the sharp wall around the solution of SGD. The test MCR curve is also shown in this figure. Surprisingly, the test MCR obtained by S-SGD, which is 30.52\%, is much lower than that of SGD, 31.74\%. This is because the statistics of test data are not exactly the same as those of the training data. This figure clearly shows the regularization of S-SGD and its effects on generalization. The two-dimensional version of this plot is shown in Section \ref{sec:discussion}.






The amount of computation demanded for S-SGD is about twice that of the conventional SGD. The convergence speed and reduction of computation are discussed in Section \ref{sec:optimize}. 
 

\section{Optimization of S-SGD hyperparameters}\label{sec:optimize}
Hyperparameters need to be carefully selected for DNN training. In this section, we consider the effects of injected noise levels and those of injection. We also perform an ablation study and propose a few variations of the S-SGD method that do not demand increased amount of training time. The large batch training results are also presented. 
 
We conducted experiments with ResNet20, a popular model that employs batch normalization \cite{ioffe2015batch}. CIFAR-10 and CIFAR-100 \cite{krizhevsky2010cifar} datasets were used for parameter optimization. We applied channel-wise normalization to the training data and augmented them by random cropping with a size of 4 and horizontal flipping following \cite{lee2015deeply}. A momentum optimizer with a batch size of 128 was used for training, unless otherwise specified. The initial learning rate was 0.1, which decays by a factor of 10 at epochs of 75 and 125. We trained the models for 175 epochs. All the experiments were implemented in TensorFlow \cite{abadi2016tensorflow}.

\begin{figure}[t]
\centering
\begin{tikzpicture}
\begin{axis}[
    xlabel={Noise level [$\sigma$]},
    ylabel={Accuracy[\%]},
    xmin=0, xmax=1,
    ymin=62, ymax=70,
    height=0.65\linewidth,
    width=1.0\linewidth,
    xtick={0,0.2,0.4,0.6,0.8,1.0},
    ytick={62,63,64,65,66,67,68,69,70},
    legend pos=south west,
    legend style={font=\fontsize{6}{7}\selectfont},
    ymajorgrids=true,
    grid style=dashed,
]
    
\addplot[
    color=red,
    mark=square,
    ]
    coordinates {
    (0   ,67.99)
    (	0.1	,	68.334	)
(	0.2	,	68.624	)
(	0.3	,	69.196	)
(	0.4	,	69.416	)
(	0.5	,	69.398	)
(	0.6	,	68.792	)
(	0.7	,	68.296	)
(	0.8	,	67.108	)
(	0.9	,	65.4	)
(	1	,	64.526	)
    };
    \addlegendentry{S-SGD}

\addplot[
    color=blue,
    mark=square,
    ]
    coordinates {
    (0   ,67.99)
    (	0.1	,	68.425	)
(	0.2	,	68.725	)
(	0.3	,	69.12	)
(	0.4	,	69.32	)
(	0.5	,	69.28	)
(	0.6	,	69.1	)
(	0.7	,	68.3	)
(	0.8	,	67.72	)
(	0.9	,	66.68	)
(	1	,	65.36	)

    };
    \addlegendentry{S-SGD Conv only}
\addplot[
    color=green,
    mark=square,
    ]
    coordinates {
    (0   ,67.99)(	0.1	,	68.18	)
(	0.2	,	68.16	)
(	0.3	,	68.22	)
(	0.4	,	68.18	)
(	0.5	,	67.88	)
(	0.6	,	67.8	)
(	0.7	,	68.18	)
(	0.8	,	68.12	)
(	0.9	,	67.68	)
(	1	,	67.72	)

    };
    \addlegendentry{S-SGD Dense only}
\addplot[
    color=Sepia,
    mark=square,
    ]
    coordinates {
    (0   ,67.99)
    (	0.1	,	68.118	)
(	0.2	,	68.308	)
(	0.3	,	68.694	)
(	0.4	,	68.72	)
(	0.5	,	68.264	)
(	0.6	,	67.738	)
(	0.7	,	66.662	)
(	0.8	,	65.92	)
(	0.9	,	64.79	)
(	1	,	62.878	)
    };
    \addlegendentry{SmoothOut}
\addplot[dashed]coordinates{(0,67.4)(1,67.4)};\addlegendentry{SGD+Dropout(0.2)}
\addplot[name path=convOnly_p, draw=none] coordinates{
(0  ,   68.38) (	0.1	,	68.77571356	)
(	0.2	,	69.12873258	)
(	0.3	,	69.38832816	)
(	0.4	,	69.66205263	)
(	0.5	,	69.42832397	)
(	0.6	,	69.22247449	)
(	0.7	,	68.62403703	)
(	0.8	,	67.97884358	)
(	0.9	,	67.05013511	)
(	1	,	65.63018512	)

};
\addplot[name path=convOnly_m, draw=none] coordinates{
(0  ,   67.6) (	0.1	,	68.07428644	)
(	0.2	,	68.32126742	)
(	0.3	,	68.85167184	)
(	0.4	,	68.97794737	)
(	0.5	,	69.13167603	)
(	0.6	,	68.97752551	)
(	0.7	,	67.97596297	)
(	0.8	,	67.46115642	)
(	0.9	,	66.30986489	)
(	1	,	65.08981488	)

};
\addplot[color=blue, opacity=0.2] fill between[of=convOnly_p and convOnly_m];
\addplot[name path=denseOnly_p, draw=none] coordinates{
(0  ,   68.38)(	0.1	,	68.53637059	)
(	0.2	,	68.56373258	)
(	0.3	,	68.70166378	)
(	0.4	,	68.35888544	)
(	0.5	,	68.19937439	)
(	0.6	,	68.18729833	)
(	0.7	,	68.46635642	)
(	0.8	,	68.51623226	)
(	0.9	,	68.02205263	)
(	1	,	68.29183914	)
};
\addplot[name path=denseOnly_m, draw=none] coordinates{
(0  ,   67.6)(	0.1	,	67.82362941	)
(	0.2	,	67.75626742	)
(	0.3	,	67.73833622	)
(	0.4	,	68.00111456	)
(	0.5	,	67.56062561	)
(	0.6	,	67.41270167	)
(	0.7	,	67.89364358	)
(	0.8	,	67.72376774	)
(	0.9	,	67.33794737	)
(	1	,	67.14816086	)
};
\addplot[color=green, opacity=0.2] fill between[of=denseOnly_p and denseOnly_m];
\addplot[name path=ssgd_p, draw=none] coordinates{
(0  ,   68.38)
(	0.1	,	68.53655863	)
(	0.2	,	68.88919804	)
(	0.3	,	69.51100794	)
(	0.4	,	69.66515858	)
(	0.5	,	69.63542367	)
(	0.6	,	69.09211664	)
(	0.7	,	68.66018402	)
(	0.8	,	67.60086915	)
(	0.9	,	66.11940948	)
(	1	,	65.06542562	)
};
\addplot[name path=ssgd_n, draw=none] coordinates{
(0  ,   67.6)(	0.1	,	68.13144137	)(	0.2	,	68.35880196	)(	0.3	,	68.88099206	)(	0.4	,	69.16684142	)(	0.5	,	69.16057633	)(	0.6	,	68.49188336	)(	0.7	,	67.93181598	)(	0.8	,	66.61513085	)(	0.9	,	64.68059052	)(	1	,	63.98657438	)
};\addplot[color=red, opacity=0.2] fill between[of=ssgd_p and ssgd_n];
\addplot[name path=smooth_p, draw=none] coordinates{
(0  ,   68.38)(	0.1	,	68.32310973	)
(	0.2	,	68.66257016	)
(	0.3	,	69.17115825	)
(	0.4	,	69.12981703	)
(	0.5	,	68.58885381	)
(	0.6	,	68.16411031	)
(	0.7	,	67.18182689	)
(	0.8	,	66.42857644	)
(	0.9	,	65.20731283	)
(	1	,	63.61009972	)
};
\addplot[name path=smooth_n, draw=none] coordinates{
(0  ,   67.6)(	0.1	,	67.91289027	)
(	0.2	,	67.95342984	)
(	0.3	,	68.21684175	)
(	0.4	,	68.31018297	)
(	0.5	,	67.93914619	)
(	0.6	,	67.31188969	)
(	0.7	,	66.14217311	)
(	0.8	,	65.41142356	)
(	0.9	,	64.37268717	)
(	1	,	62.14590028	)
};\addplot[color=Sepia, opacity=0.2] fill between[of=smooth_p and smooth_n ];
    \end{axis}
\end{tikzpicture}
\caption{Test accuracy according to the amount of injected noise in S-SGD and SmoothOut. Average test accuracy value and its error bar are computed from 5 runs. S-SGD Conv only and S-SGD Dense only denotes the results of the selective noise injection. Dropout is only applied to a dense layer in SGD+Dropout.}
\label{fig:fig_bell_curve}  
\end{figure}
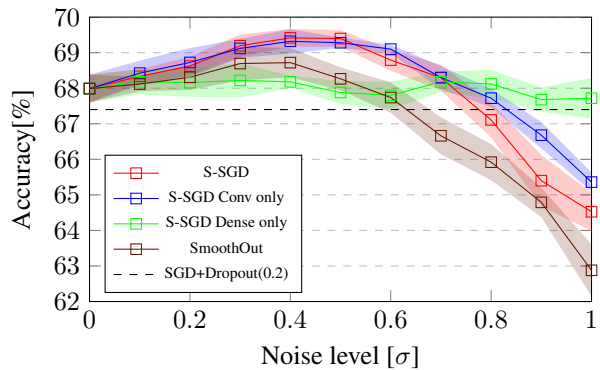

\subsection{Optimum Noise Level}\label{sec:optim_noise}

The performance of S-SGD depends on the amount of injected noises. We also observed the effects of selective noise injection to either convolution or dense layers. Note that the Dropout and DropConnect were developed for the regularization of parameter rich dense layers. 

We observed the effect of noise level for S-SGD and SmoothOut with ResNet20 on CIFAR-100 dataset. The SmoothOut is one of the weight noise injection algorithms \cite{wen2018smoothout}. The noises were injected into weights of convolution and dense layers. \figurename~\ref{fig:fig_bell_curve} shows that both S-SGD and SmoothOut yield the best results when the noise level is 0.4 $\sigma$ of the weights, where $\sigma$ is the L2 norm of weights for each layer. Although the optimum level is similar, S-SGD injects stronger noise than SmoothOut because it injects two noises and the noise strength of the former is always constant, while that of the latter is uniformly distributed between 0 and the specified noise level. Note that S-SGD shows consistent improvement at broad noise levels. \figurename~\ref{fig:fig_bell_curve} also contains the results when the symmetrical noises are injected only to convolution or dense layers of ResNet20. We can observe that the symmetrical noise injection to convolution layers also yields good results, indicating that S-SGD is a flexible regularization method when compared to Dropout or DropConnect.

\subsection{Convergence Curve and Training Time Reduction}

The S-SGD method employs two forward and backward passes; thus, the training time for each epoch increases when compared to that of the conventional SGD method. Although this work was intended to find the best model, we examined the convergence speed. 

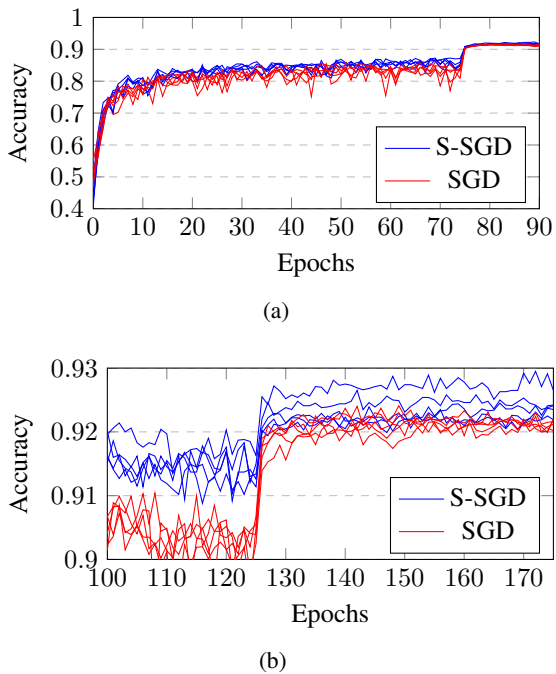
\begin{figure}[h]
\centering
\subfigure[]{
\centering
\begin{tikzpicture}
\begin{axis}[
    xlabel={Epochs},
    ylabel={Accuracy},
    xmin=0, xmax=90,
    ymin=0.40, ymax=1.00,
    width = 0.91\linewidth,
    height = 0.5\linewidth,
    xtick={0,10,20,30,40,50,60,70,80,90},
    ytick={0.40,0.50,0.60,0.70,0.80,0.90,1.00},
    legend pos=south east,
    ymajorgrids=true,
    grid style=dashed,
]
\addplot[
    color=blue
    ]
    coordinates {
    (0, 0.4461)
    (1, 0.626)
    (2, 0.6945999)
    (3, 0.72010005)
    (4, 0.7647)
    (5, 0.7666)
    (6, 0.7678)
    (7, 0.8063)
    (8, 0.7958)
    (9, 0.7986)
    (10, 0.8112)
    (11, 0.7544)
    (12, 0.78449994)
    (13, 0.8333)
    (14, 0.7983)
    (15, 0.81659997)
    (16, 0.83980006)
    (17, 0.8301)
    (18, 0.8281999)
    (19, 0.8281)
    (20, 0.7987999)
    (21, 0.80189985)
    (22, 0.84090006)
    (23, 0.83409995)
    (24, 0.8279)
    (25, 0.8215)
    (26, 0.829)
    (27, 0.83839995)
    (28, 0.8434999)
    (29, 0.84560007)
    (30, 0.8535)
    (31, 0.84480006)
    (32, 0.8557)
    (33, 0.8613)
    (34, 0.8458001)
    (35, 0.8301)
    (36, 0.84059995)
    (37, 0.8535999)
    (38, 0.84879994)
    (39, 0.8504999)
    (40, 0.84840006)
    (41, 0.8507999)
    (42, 0.8468999)
    (43, 0.84760004)
    (44, 0.8543)
    (45, 0.8514)
    (46, 0.85020006)
    (47, 0.8169)
    (48, 0.81)
    (49, 0.8503001)
    (50, 0.8426001)
    (51, 0.84139997)
    (52, 0.86)
    (53, 0.8397001)
    (54, 0.8660001)
    (55, 0.84349996)
    (56, 0.83770007)
    (57, 0.86060005)
    (58, 0.84300005)
    (59, 0.87)
    (60, 0.8629001)
    (61, 0.8618)
    (62, 0.8566001)
    (63, 0.8370999)
    (64, 0.86179996)
    (65, 0.8714)
    (66, 0.8645)
    (67, 0.8256999)
    (68, 0.8699)
    (69, 0.87240005)
    (70, 0.8522001)
    (71, 0.87060004)
    (72, 0.87249994)
    (73, 0.87000006)
    (74, 0.87090003)
    (75, 0.90870005)
    (76, 0.9128)
    (77, 0.9174)
    (78, 0.9164001)
    (79, 0.9198)
    (80, 0.9193)
    (81, 0.9187)
    (82, 0.91829985)
    (83, 0.92149997)
    (84, 0.91979986)
    (85, 0.9189001)
    (86, 0.9189)
    (87, 0.92170006)
    (88, 0.92159986)
    (89, 0.923)
    (90, 0.9175)

    };
    \addplot[
    color=red
    ]
    coordinates {
    (0, 0.5504)
    (1, 0.62530005)
    (2, 0.7124001)
    (3, 0.7349)
    (4, 0.69539994)
    (5, 0.7019)
    (6, 0.73859996)
    (7, 0.7619)
    (8, 0.75179994)
    (9, 0.7804001)
    (10, 0.7881999)
    (11, 0.7761)
    (12, 0.7299999)
    (13, 0.8091999)
    (14, 0.8174)
    (15, 0.7944)
    (16, 0.8108)
    (17, 0.8111)
    (18, 0.795)
    (19, 0.79280007)
    (20, 0.8163)
    (21, 0.81240004)
    (22, 0.83129996)
    (23, 0.82080007)
    (24, 0.80560005)
    (25, 0.8265)
    (26, 0.8226)
    (27, 0.786)
    (28, 0.82710004)
    (29, 0.83709997)
    (30, 0.8375001)
    (31, 0.8015)
    (32, 0.8201)
    (33, 0.83519995)
    (34, 0.82179993)
    (35, 0.78089994)
    (36, 0.8291)
    (37, 0.828)
    (38, 0.83940005)
    (39, 0.8207)
    (40, 0.83910006)
    (41, 0.78800005)
    (42, 0.82430005)
    (43, 0.8391001)
    (44, 0.75209993)
    (45, 0.84760004)
    (46, 0.8135998)
    (47, 0.84809995)
    (48, 0.83079994)
    (49, 0.826)
    (50, 0.8459999)
    (51, 0.83009994)
    (52, 0.81810004)
    (53, 0.82739997)
    (54, 0.84549993)
    (55, 0.833)
    (56, 0.84159994)
    (57, 0.84540004)
    (58, 0.8265)
    (59, 0.834)
    (60, 0.84139997)
    (61, 0.8134001)
    (62, 0.83089995)
    (63, 0.84090006)
    (64, 0.84379995)
    (65, 0.8338999)
    (66, 0.86320007)
    (67, 0.83729994)
    (68, 0.8338001)
    (69, 0.86000013)
    (70, 0.84519994)
    (71, 0.84290004)
    (72, 0.8279)
    (73, 0.8379999)
    (74, 0.8318)
    (75, 0.90449995)
    (76, 0.9104001)
    (77, 0.9115999)
    (78, 0.9146)
    (79, 0.91409993)
    (80, 0.9153001)
    (81, 0.91260004)
    (82, 0.9134999)
    (83, 0.91159993)
    (84, 0.9128)
    (85, 0.9115001)
    (86, 0.91269994)
    (87, 0.91159993)
    (88, 0.9129001)
    (89, 0.91200006)
    (90, 0.9118999)

    };
\addplot[
    color=blue
    ]
    coordinates {
    (0, 0.408)
    (1, 0.6009)
    (2, 0.6682001)
    (3, 0.72220004)
    (4, 0.7058)
    (5, 0.77380013)
    (6, 0.75570005)
    (7, 0.81020004)
    (8, 0.79299986)
    (9, 0.7942001)
    (10, 0.77669996)
    (11, 0.7652)
    (12, 0.7885)
    (13, 0.81560016)
    (14, 0.8047)
    (15, 0.81860006)
    (16, 0.8134)
    (17, 0.8077)
    (18, 0.8282)
    (19, 0.8318)
    (20, 0.8378001)
    (21, 0.7962001)
    (22, 0.8511999)
    (23, 0.80920005)
    (24, 0.8255)
    (25, 0.8493001)
    (26, 0.83659995)
    (27, 0.82989997)
    (28, 0.8450999)
    (29, 0.8543)
    (30, 0.8536999)
    (31, 0.8449001)
    (32, 0.81939995)
    (33, 0.84699994)
    (34, 0.8443999)
    (35, 0.8354001)
    (36, 0.84050006)
    (37, 0.85240006)
    (38, 0.8527)
    (39, 0.83519995)
    (40, 0.8274)
    (41, 0.84729993)
    (42, 0.85330003)
    (43, 0.8522001)
    (44, 0.84900004)
    (45, 0.8528)
    (46, 0.8443001)
    (47, 0.8303001)
    (48, 0.8173)
    (49, 0.86589986)
    (50, 0.84700006)
    (51, 0.8484001)
    (52, 0.8523999)
    (53, 0.84810007)
    (54, 0.8615)
    (55, 0.84760004)
    (56, 0.84799993)
    (57, 0.86850005)
    (58, 0.85840005)
    (59, 0.8514)
    (60, 0.85159993)
    (61, 0.8618)
    (62, 0.85190004)
    (63, 0.8592)
    (64, 0.85260004)
    (65, 0.8608)
    (66, 0.85150003)
    (67, 0.8595)
    (68, 0.8636)
    (69, 0.86420006)
    (70, 0.8667)
    (71, 0.87079984)
    (72, 0.859)
    (73, 0.8391001)
    (74, 0.8518999)
    (75, 0.90639997)
    (76, 0.9121)
    (77, 0.9103001)
    (78, 0.9143999)
    (79, 0.9148001)
    (80, 0.916)
    (81, 0.916)
    (82, 0.9170999)
    (83, 0.9167)
    (84, 0.91700006)
    (85, 0.9167001)
    (86, 0.91920006)
    (87, 0.918)
    (88, 0.9142)
    (89, 0.9177)
    (90, 0.91600007)

    };
\addplot[
    color=blue
    ]
    coordinates {
    (0, 0.4297)
    (1, 0.5749)
    (2, 0.6819)
    (3, 0.74740005)
    (4, 0.7012)
    (5, 0.78510004)
    (6, 0.7473)
    (7, 0.7843)
    (8, 0.7757)
    (9, 0.8026)
    (10, 0.78629994)
    (11, 0.7969)
    (12, 0.78029996)
    (13, 0.8142)
    (14, 0.82130015)
    (15, 0.8036)
    (16, 0.8226)
    (17, 0.8348)
    (18, 0.8213999)
    (19, 0.79690003)
    (20, 0.82559997)
    (21, 0.8277999)
    (22, 0.8386)
    (23, 0.81)
    (24, 0.82170004)
    (25, 0.8398999)
    (26, 0.84980005)
    (27, 0.82559997)
    (28, 0.8334)
    (29, 0.8286)
    (30, 0.8376)
    (31, 0.83730006)
    (32, 0.8456001)
    (33, 0.8352001)
    (34, 0.8178)
    (35, 0.8387001)
    (36, 0.8578)
    (37, 0.85450006)
    (38, 0.8518999)
    (39, 0.8340001)
    (40, 0.8315)
    (41, 0.84940004)
    (42, 0.8514)
    (43, 0.84330004)
    (44, 0.8282001)
    (45, 0.84810007)
    (46, 0.8202)
    (47, 0.8222)
    (48, 0.85020006)
    (49, 0.8596)
    (50, 0.86)
    (51, 0.85590005)
    (52, 0.8488999)
    (53, 0.85740006)
    (54, 0.86120003)
    (55, 0.82629997)
    (56, 0.8493)
    (57, 0.86849993)
    (58, 0.8678)
    (59, 0.84610003)
    (60, 0.8596)
    (61, 0.8474999)
    (62, 0.85730004)
    (63, 0.86509997)
    (64, 0.8525)
    (65, 0.8605)
    (66, 0.85870004)
    (67, 0.84559995)
    (68, 0.8489)
    (69, 0.86820006)
    (70, 0.8549999)
    (71, 0.8648)
    (72, 0.8536999)
    (73, 0.8374999)
    (74, 0.84730005)
    (75, 0.90880007)
    (76, 0.91080004)
    (77, 0.91220003)
    (78, 0.91420007)
    (79, 0.9143)
    (80, 0.91760004)
    (81, 0.9159)
    (82, 0.9157)
    (83, 0.91800004)
    (84, 0.9182)
    (85, 0.9157001)
    (86, 0.91520005)
    (87, 0.9145)
    (88, 0.9150001)
    (89, 0.91730005)
    (90, 0.9151001)

    };
\addplot[
    color=blue
    ]
    coordinates {
    (0, 0.49409997)
    (1, 0.6452)
    (2, 0.7329)
    (3, 0.7476)
    (4, 0.751)
    (5, 0.79100007)
    (6, 0.80419993)
    (7, 0.8184999)
    (8, 0.79710007)
    (9, 0.8093)
    (10, 0.80880004)
    (11, 0.7860999)
    (12, 0.8141)
    (13, 0.83739996)
    (14, 0.82859993)
    (15, 0.82089996)
    (16, 0.8321)
    (17, 0.8286001)
    (18, 0.82629997)
    (19, 0.8289)
    (20, 0.83449996)
    (21, 0.826)
    (22, 0.8456999)
    (23, 0.8269)
    (24, 0.84019995)
    (25, 0.84909993)
    (26, 0.8270999)
    (27, 0.8115)
    (28, 0.8417999)
    (29, 0.8386001)
    (30, 0.84479994)
    (31, 0.8503)
    (32, 0.8379)
    (33, 0.85370004)
    (34, 0.84619987)
    (35, 0.8345001)
    (36, 0.8528001)
    (37, 0.8390001)
    (38, 0.8509001)
    (39, 0.81949997)
    (40, 0.8428001)
    (41, 0.84840006)
    (42, 0.84330004)
    (43, 0.8418)
    (44, 0.81869996)
    (45, 0.8529001)
    (46, 0.85400003)
    (47, 0.8205)
    (48, 0.84319997)
    (49, 0.8627999)
    (50, 0.8683)
    (51, 0.857)
    (52, 0.8456001)
    (53, 0.8276)
    (54, 0.8583)
    (55, 0.8511)
    (56, 0.8549999)
    (57, 0.8599999)
    (58, 0.84059995)
    (59, 0.85590005)
    (60, 0.8349999)
    (61, 0.8417001)
    (62, 0.8633001)
    (63, 0.84449995)
    (64, 0.8415001)
    (65, 0.8410001)
    (66, 0.8614)
    (67, 0.859)
    (68, 0.8709002)
    (69, 0.86310005)
    (70, 0.8671)
    (71, 0.8515001)
    (72, 0.8642)
    (73, 0.8521)
    (74, 0.85940003)
    (75, 0.9075001)
    (76, 0.9094999)
    (77, 0.9131001)
    (78, 0.91600007)
    (79, 0.9154)
    (80, 0.9162999)
    (81, 0.91560006)
    (82, 0.91730005)
    (83, 0.9166999)
    (84, 0.9173)
    (85, 0.91780007)
    (86, 0.9205)
    (87, 0.9175999)
    (88, 0.9194)
    (89, 0.9154)
    (90, 0.9146001)

    };
\addplot[
    color=blue
    ]
    coordinates {
    (0, 0.42400002)
    (1, 0.59529996)
    (2, 0.7075)
    (3, 0.73200005)
    (4, 0.7709)
    (5, 0.7739001)
    (6, 0.7883999)
    (7, 0.8097)
    (8, 0.7904999)
    (9, 0.8011)
    (10, 0.7594999)
    (11, 0.76700014)
    (12, 0.7928)
    (13, 0.8298)
    (14, 0.7864)
    (15, 0.8155)
    (16, 0.8245999)
    (17, 0.81049997)
    (18, 0.82340014)
    (19, 0.8091)
    (20, 0.8287)
    (21, 0.79969996)
    (22, 0.84220004)
    (23, 0.8237)
    (24, 0.8361)
    (25, 0.84)
    (26, 0.8157)
    (27, 0.8453001)
    (28, 0.84160006)
    (29, 0.8351001)
    (30, 0.8432999)
    (31, 0.80520004)
    (32, 0.8246)
    (33, 0.8457001)
    (34, 0.84169996)
    (35, 0.8142)
    (36, 0.84540004)
    (37, 0.85590005)
    (38, 0.85600007)
    (39, 0.83529997)
    (40, 0.8468)
    (41, 0.8458001)
    (42, 0.8553)
    (43, 0.84019995)
    (44, 0.8437999)
    (45, 0.8182)
    (46, 0.8645)
    (47, 0.83699995)
    (48, 0.8379001)
    (49, 0.8543999)
    (50, 0.8602)
    (51, 0.8524)
    (52, 0.86219996)
    (53, 0.85130006)
    (54, 0.8458999)
    (55, 0.8405001)
    (56, 0.83709997)
    (57, 0.8532)
    (58, 0.8539001)
    (59, 0.8555)
    (60, 0.85980004)
    (61, 0.8492)
    (62, 0.8588)
    (63, 0.8573)
    (64, 0.8675)
    (65, 0.85190004)
    (66, 0.8656001)
    (67, 0.84699994)
    (68, 0.86730003)
    (69, 0.86579996)
    (70, 0.8428001)
    (71, 0.8339001)
    (72, 0.8613)
    (73, 0.84119993)
    (74, 0.8527)
    (75, 0.9087001)
    (76, 0.9102999)
    (77, 0.9139)
    (78, 0.9157001)
    (79, 0.9132999)
    (80, 0.9147999)
    (81, 0.91590005)
    (82, 0.9157)
    (83, 0.9177)
    (84, 0.9150999)
    (85, 0.9163)
    (86, 0.9184)
    (87, 0.9177)
    (88, 0.918)
    (89, 0.91619986)
    (90, 0.918)

    };
    \addlegendentry{S-SGD}

\addplot[
    color=red
    ]
    coordinates {
    (0, 0.48729995)
    (1, 0.5721)
    (2, 0.6575)
    (3, 0.73640007)
    (4, 0.7325)
    (5, 0.7672001)
    (6, 0.76280004)
    (7, 0.791)
    (8, 0.763)
    (9, 0.7888999)
    (10, 0.7745001)
    (11, 0.78119993)
    (12, 0.77639997)
    (13, 0.81640005)
    (14, 0.80590004)
    (15, 0.8151)
    (16, 0.79020005)
    (17, 0.821)
    (18, 0.8237)
    (19, 0.79649997)
    (20, 0.8266999)
    (21, 0.79620004)
    (22, 0.8172)
    (23, 0.77369994)
    (24, 0.81640005)
    (25, 0.81560004)
    (26, 0.7643001)
    (27, 0.8164)
    (28, 0.83559996)
    (29, 0.8077001)
    (30, 0.8034)
    (31, 0.81289995)
    (32, 0.8198)
    (33, 0.82850003)
    (34, 0.81069994)
    (35, 0.81450003)
    (36, 0.8307)
    (37, 0.83439994)
    (38, 0.8381001)
    (39, 0.8157002)
    (40, 0.8325)
    (41, 0.8249)
    (42, 0.8109)
    (43, 0.7944001)
    (44, 0.8141999)
    (45, 0.81569993)
    (46, 0.81740004)
    (47, 0.8372001)
    (48, 0.82949996)
    (49, 0.84379995)
    (50, 0.82100016)
    (51, 0.8346001)
    (52, 0.84559995)
    (53, 0.82829994)
    (54, 0.8440001)
    (55, 0.83449996)
    (56, 0.8517999)
    (57, 0.84160006)
    (58, 0.8375)
    (59, 0.8370999)
    (60, 0.8318001)
    (61, 0.7936001)
    (62, 0.8318)
    (63, 0.8386)
    (64, 0.8428999)
    (65, 0.8513)
    (66, 0.8185)
    (67, 0.82519996)
    (68, 0.8243)
    (69, 0.8396999)
    (70, 0.824)
    (71, 0.8518)
    (72, 0.7704)
    (73, 0.83059996)
    (74, 0.80930007)
    (75, 0.90489995)
    (76, 0.90860003)
    (77, 0.91089994)
    (78, 0.91229993)
    (79, 0.91380006)
    (80, 0.9138)
    (81, 0.9161001)
    (82, 0.9158001)
    (83, 0.9174999)
    (84, 0.9177999)
    (85, 0.91539985)
    (86, 0.9160999)
    (87, 0.91829985)
    (88, 0.9162001)
    (89, 0.9103)
    (90, 0.9135999)

    };
\addplot[
    color=red
    ]
    coordinates {
    (0, 0.4687)
    (1, 0.64)
    (2, 0.69159997)
    (3, 0.7491001)
    (4, 0.73699987)
    (5, 0.7825)
    (6, 0.7463)
    (7, 0.7712)
    (8, 0.72720003)
    (9, 0.77579993)
    (10, 0.7632)
    (11, 0.78730005)
    (12, 0.8013)
    (13, 0.8057)
    (14, 0.78729993)
    (15, 0.7846001)
    (16, 0.79420006)
    (17, 0.7979001)
    (18, 0.8254001)
    (19, 0.8104)
    (20, 0.8007)
    (21, 0.75450003)
    (22, 0.7763999)
    (23, 0.81170005)
    (24, 0.7875)
    (25, 0.82680017)
    (26, 0.8047)
    (27, 0.8207)
    (28, 0.8324001)
    (29, 0.82369995)
    (30, 0.8115001)
    (31, 0.83019996)
    (32, 0.8011)
    (33, 0.8317001)
    (34, 0.80899996)
    (35, 0.80039996)
    (36, 0.81009996)
    (37, 0.83610016)
    (38, 0.82509995)
    (39, 0.7887999)
    (40, 0.8027)
    (41, 0.80899996)
    (42, 0.82729995)
    (43, 0.82789993)
    (44, 0.8052)
    (45, 0.8402001)
    (46, 0.81250006)
    (47, 0.8544001)
    (48, 0.7909)
    (49, 0.7995999)
    (50, 0.81920004)
    (51, 0.83950007)
    (52, 0.81380004)
    (53, 0.8104)
    (54, 0.84089994)
    (55, 0.8009)
    (56, 0.8485)
    (57, 0.8303001)
    (58, 0.8284)
    (59, 0.7548001)
    (60, 0.83449996)
    (61, 0.84400004)
    (62, 0.8406001)
    (63, 0.8374999)
    (64, 0.83549994)
    (65, 0.84479994)
    (66, 0.8579)
    (67, 0.8284)
    (68, 0.81490016)
    (69, 0.8355)
    (70, 0.8152)
    (71, 0.83329993)
    (72, 0.8348)
    (73, 0.8336)
    (74, 0.8358001)
    (75, 0.90189993)
    (76, 0.90799993)
    (77, 0.90979993)
    (78, 0.9119998)
    (79, 0.91229993)
    (80, 0.9144999)
    (81, 0.9153001)
    (82, 0.9127999)
    (83, 0.9146)
    (84, 0.91420007)
    (85, 0.9118999)
    (86, 0.91229993)
    (87, 0.9135999)
    (88, 0.9135001)
    (89, 0.9119999)
    (90, 0.91089994)
    };
\addplot[
    color=red
    ]
    coordinates {
    (0, 0.51320004)
    (1, 0.57640004)
    (2, 0.6504)
    (3, 0.74500006)
    (4, 0.7235)
    (5, 0.7385)
    (6, 0.7837999)
    (7, 0.77840006)
    (8, 0.77659994)
    (9, 0.7844999)
    (10, 0.7576999)
    (11, 0.8121)
    (12, 0.78649986)
    (13, 0.8054)
    (14, 0.7526)
    (15, 0.79560006)
    (16, 0.78840005)
    (17, 0.7872001)
    (18, 0.80539995)
    (19, 0.82460004)
    (20, 0.8206)
    (21, 0.7768)
    (22, 0.82369995)
    (23, 0.8091)
    (24, 0.7975)
    (25, 0.7947)
    (26, 0.83380014)
    (27, 0.81060004)
    (28, 0.8385999)
    (29, 0.80329996)
    (30, 0.82939994)
    (31, 0.8195)
    (32, 0.8176)
    (33, 0.8352999)
    (34, 0.80680007)
    (35, 0.81049997)
    (36, 0.8232)
    (37, 0.84480006)
    (38, 0.8211)
    (39, 0.80579996)
    (40, 0.79940003)
    (41, 0.8269)
    (42, 0.8195999)
    (43, 0.84809995)
    (44, 0.832)
    (45, 0.83379996)
    (46, 0.83289987)
    (47, 0.83520013)
    (48, 0.8173999)
    (49, 0.81890005)
    (50, 0.82360005)
    (51, 0.85910004)
    (52, 0.8396999)
    (53, 0.84440005)
    (54, 0.8303)
    (55, 0.84489995)
    (56, 0.8503001)
    (57, 0.83660007)
    (58, 0.8196)
    (59, 0.8288999)
    (60, 0.84560007)
    (61, 0.8072)
    (62, 0.85010004)
    (63, 0.8459001)
    (64, 0.8316)
    (65, 0.8224)
    (66, 0.84370005)
    (67, 0.8323)
    (68, 0.8545999)
    (69, 0.84589994)
    (70, 0.78510004)
    (71, 0.8321)
    (72, 0.7977)
    (73, 0.81020004)
    (74, 0.839)
    (75, 0.9050999)
    (76, 0.90709996)
    (77, 0.91290003)
    (78, 0.91280013)
    (79, 0.91400003)
    (80, 0.9142)
    (81, 0.91339993)
    (82, 0.9136)
    (83, 0.9150999)
    (84, 0.9128)
    (85, 0.9137001)
    (86, 0.91220003)
    (87, 0.91200006)
    (88, 0.9114001)
    (89, 0.90840006)
    (90, 0.90989995)

    };
\addplot[
    color=red
    ]
    coordinates {
    (0, 0.5326)
    (1, 0.56079996)
    (2, 0.6992)
    (3, 0.72190005)
    (4, 0.75819993)
    (5, 0.7455999)
    (6, 0.7745001)
    (7, 0.7712999)
    (8, 0.7584999)
    (9, 0.74280006)
    (10, 0.8057)
    (11, 0.7907)
    (12, 0.7973001)
    (13, 0.80439997)
    (14, 0.81240004)
    (15, 0.82)
    (16, 0.81839997)
    (17, 0.8208)
    (18, 0.7976001)
    (19, 0.8328999)
    (20, 0.8289)
    (21, 0.8069999)
    (22, 0.8057)
    (23, 0.8148)
    (24, 0.8174)
    (25, 0.78839993)
    (26, 0.8207)
    (27, 0.7672)
    (28, 0.8122)
    (29, 0.83119994)
    (30, 0.8274)
    (31, 0.8168)
    (32, 0.8294)
    (33, 0.84529996)
    (34, 0.8497001)
    (35, 0.8122)
    (36, 0.8301)
    (37, 0.8434001)
    (38, 0.84220016)
    (39, 0.84730005)
    (40, 0.8274)
    (41, 0.8364)
    (42, 0.8332)
    (43, 0.82009995)
    (44, 0.84200007)
    (45, 0.8375)
    (46, 0.8557)
    (47, 0.85089993)
    (48, 0.8465001)
    (49, 0.8525999)
    (50, 0.8329)
    (51, 0.84139997)
    (52, 0.80350006)
    (53, 0.84970003)
    (54, 0.8550999)
    (55, 0.8493)
    (56, 0.8492)
    (57, 0.82629997)
    (58, 0.83989996)
    (59, 0.8091999)
    (60, 0.83799994)
    (61, 0.7886)
    (62, 0.83840007)
    (63, 0.8339001)
    (64, 0.82810014)
    (65, 0.82879996)
    (66, 0.8432001)
    (67, 0.84299994)
    (68, 0.8273999)
    (69, 0.8485001)
    (70, 0.84690005)
    (71, 0.84330004)
    (72, 0.8522999)
    (73, 0.83559996)
    (74, 0.83909994)
    (75, 0.9028001)
    (76, 0.9096001)
    (77, 0.9110001)
    (78, 0.91420007)
    (79, 0.9117)
    (80, 0.9156)
    (81, 0.9154001)
    (82, 0.9146)
    (83, 0.91409993)
    (84, 0.9134999)
    (85, 0.91460013)
    (86, 0.9154)
    (87, 0.91339993)
    (88, 0.91190004)
    (89, 0.91260004)
    (90, 0.91719985)
    };
    \addlegendentry{SGD}
 
\end{axis}
\end{tikzpicture}
}
\subfigure[]{
\centering
\begin{tikzpicture}
\begin{axis}[
    xlabel={Epochs},
    ylabel={Accuracy},
    xmin=100, xmax=175,
    ymin=0.90, ymax=0.93,
    width = 0.91\linewidth,
    height = 0.5\linewidth,
    xtick={100,110,120,130,140,150,160,170},
    ytick={0.90,0.91, 0.92, 0.93},
    legend pos=south east,
    ymajorgrids=true,
    grid style=dashed,
]
\addplot[
    color=blue
    ]
    coordinates {
    (100, 0.9196)
    (101, 0.9206)
    (102, 0.9214)
    (103, 0.9200001)
    (104, 0.92060006)
    (105, 0.91829985)
    (106, 0.9175999)
    (107, 0.91830003)
    (108, 0.91800004)
    (109, 0.9204)
    (110, 0.9197)
    (111, 0.9135001)
    (112, 0.9154)
    (113, 0.9164999)
    (114, 0.9171999)
    (115, 0.9173999)
    (116, 0.91659987)
    (117, 0.919)
    (118, 0.91580003)
    (119, 0.91650003)
    (120, 0.92030007)
    (121, 0.91650003)
    (122, 0.9186998)
    (123, 0.91880006)
    (124, 0.9185)
    (125, 0.9163)
    (126, 0.9255)
    (127, 0.9243)
    (128, 0.9272)
    (129, 0.92639995)
    (130, 0.9252)
    (131, 0.92579997)
    (132, 0.92539996)
    (133, 0.92609996)
    (134, 0.9255999)
    (135, 0.9271)
    (136, 0.92700005)
    (137, 0.926)
    (138, 0.9280999)
    (139, 0.92729986)
    (140, 0.92749995)
    (141, 0.9273)
    (142, 0.9269)
    (143, 0.9266)
    (144, 0.92679995)
    (145, 0.9271)
    (146, 0.9273)
    (147, 0.9267)
    (148, 0.9287)
    (149, 0.9266)
    (150, 0.9274)
    (151, 0.92740005)
    (152, 0.92709994)
    (153, 0.9268)
    (154, 0.9288998)
    (155, 0.9262)
    (156, 0.9284)
    (157, 0.92679995)
    (158, 0.92759997)
    (159, 0.9284998)
    (160, 0.92780006)
    (161, 0.9266998)
    (162, 0.92699987)
    (163, 0.9277)
    (164, 0.92689997)
    (165, 0.92509997)
    (166, 0.9263)
    (167, 0.92629987)
    (168, 0.9265)
    (169, 0.92649996)
    (170, 0.92919993)
    (171, 0.9276)
    (172, 0.92949986)
    (173, 0.92709994)
    (174, 0.9291)
    (175, 0.92649996)

    };
    \addplot[
    color=red
    ]
    coordinates {
    (100, 0.90410006)
    (101, 0.90799993)
    (102, 0.90990007)
    (103, 0.90800005)
    (104, 0.9056001)
    (105, 0.90400004)
    (106, 0.90689987)
    (107, 0.9017001)
    (108, 0.90260005)
    (109, 0.90020007)
    (110, 0.9022)
    (111, 0.90450007)
    (112, 0.9029001)
    (113, 0.90369993)
    (114, 0.9055999)
    (115, 0.90330005)
    (116, 0.9031999)
    (117, 0.90089995)
    (118, 0.90440005)
    (119, 0.9057999)
    (120, 0.90620005)
    (121, 0.8977001)
    (122, 0.9036)
    (123, 0.90459996)
    (124, 0.89779997)
    (125, 0.9025001)
    (126, 0.9139)
    (127, 0.9153)
    (128, 0.9159)
    (129, 0.9179)
    (130, 0.91560006)
    (131, 0.91810006)
    (132, 0.91939986)
    (133, 0.9192)
    (134, 0.9193)
    (135, 0.91980004)
    (136, 0.9191)
    (137, 0.9197)
    (138, 0.9198)
    (139, 0.92090005)
    (140, 0.92)
    (141, 0.92010003)
    (142, 0.9206)
    (143, 0.918)
    (144, 0.9193)
    (145, 0.9198)
    (146, 0.9189999)
    (147, 0.91840005)
    (148, 0.9173999)
    (149, 0.91850007)
    (150, 0.9189001)
    (151, 0.9191999)
    (152, 0.92050004)
    (153, 0.9212)
    (154, 0.92)
    (155, 0.9215)
    (156, 0.9198)
    (157, 0.9207)
    (158, 0.9195)
    (159, 0.9203)
    (160, 0.92090005)
    (161, 0.92130005)
    (162, 0.9190001)
    (163, 0.91960007)
    (164, 0.91950005)
    (165, 0.92030007)
    (166, 0.92090005)
    (167, 0.9196)
    (168, 0.9204)
    (169, 0.9199)
    (170, 0.92160004)
    (171, 0.92010003)
    (172, 0.92080003)
    (173, 0.9211)
    (174, 0.9206)
    (175, 0.92020005)
    };
\addplot[
    color=blue
    ]
    coordinates {
    (100, 0.9152001)
    (101, 0.9186)
    (102, 0.91550004)
    (103, 0.9142999)
    (104, 0.91369987)
    (105, 0.91150004)
    (106, 0.9142999)
    (107, 0.9161999)
    (108, 0.91420007)
    (109, 0.91340005)
    (110, 0.91549987)
    (111, 0.9143)
    (112, 0.9166001)
    (113, 0.9159)
    (114, 0.91369987)
    (115, 0.9136999)
    (116, 0.9167999)
    (117, 0.9163999)
    (118, 0.9142999)
    (119, 0.91409993)
    (120, 0.9168)
    (121, 0.9140999)
    (122, 0.9172)
    (123, 0.91830003)
    (124, 0.9120999)
    (125, 0.915)
    (126, 0.92200005)
    (127, 0.92460006)
    (128, 0.92260003)
    (129, 0.9231999)
    (130, 0.9238)
    (131, 0.9234999)
    (132, 0.9248)
    (133, 0.9241)
    (134, 0.92409986)
    (135, 0.92450005)
    (136, 0.92380005)
    (137, 0.9239)
    (138, 0.9241)
    (139, 0.92439985)
    (140, 0.9244)
    (141, 0.9248)
    (142, 0.92649996)
    (143, 0.9235)
    (144, 0.9246)
    (145, 0.9238999)
    (146, 0.9246)
    (147, 0.9256)
    (148, 0.925)
    (149, 0.92490005)
    (150, 0.92310005)
    (151, 0.924)
    (152, 0.9245999)
    (153, 0.9254)
    (154, 0.92539984)
    (155, 0.92499995)
    (156, 0.9241)
    (157, 0.9258)
    (158, 0.92480004)
    (159, 0.9249)
    (160, 0.92460006)
    (161, 0.9246)
    (162, 0.92579997)
    (163, 0.92449987)
    (164, 0.9248)
    (165, 0.9236)
    (166, 0.92499995)
    (167, 0.9242)
    (168, 0.9241)
    (169, 0.92469984)
    (170, 0.9252999)
    (171, 0.92449987)
    (172, 0.9241998)
    (173, 0.9235)
    (174, 0.9238)
    (175, 0.92410004)

    };
\addplot[
    color=blue
    ]
    coordinates {
    (100, 0.9159999)
    (101, 0.9143)
    (102, 0.9164001)
    (103, 0.9130999)
    (104, 0.9146)
    (105, 0.91300005)
    (106, 0.91520005)
    (107, 0.9161999)
    (108, 0.91310006)
    (109, 0.91620004)
    (110, 0.9145)
    (111, 0.91370004)
    (112, 0.9153)
    (113, 0.9168001)
    (114, 0.9150999)
    (115, 0.91270006)
    (116, 0.90880007)
    (117, 0.9122001)
    (118, 0.9128001)
    (119, 0.9159999)
    (120, 0.91760004)
    (121, 0.9103)
    (122, 0.9132001)
    (123, 0.9178)
    (124, 0.91300005)
    (125, 0.91340005)
    (126, 0.9206)
    (127, 0.92219996)
    (128, 0.92289996)
    (129, 0.92280006)
    (130, 0.92140007)
    (131, 0.9223)
    (132, 0.92269987)
    (133, 0.9225)
    (134, 0.92260003)
    (135, 0.92200005)
    (136, 0.9229001)
    (137, 0.9219001)
    (138, 0.92280006)
    (139, 0.9232)
    (140, 0.92170006)
    (141, 0.9216999)
    (142, 0.92219996)
    (143, 0.9218)
    (144, 0.9238999)
    (145, 0.9223)
    (146, 0.9223)
    (147, 0.9216)
    (148, 0.9225001)
    (149, 0.9229)
    (150, 0.9219999)
    (151, 0.92200005)
    (152, 0.9229)
    (153, 0.9216999)
    (154, 0.92299986)
    (155, 0.9223)
    (156, 0.92329997)
    (157, 0.92339987)
    (158, 0.9243)
    (159, 0.92439985)
    (160, 0.92359996)
    (161, 0.9222)
    (162, 0.9227999)
    (163, 0.92359996)
    (164, 0.92329997)
    (165, 0.92340004)
    (166, 0.9245)
    (167, 0.9236)
    (168, 0.9236)
    (169, 0.9241)
    (170, 0.92369986)
    (171, 0.9229)
    (172, 0.92409986)
    (173, 0.9226)
    (174, 0.92409986)
    (175, 0.9241999)

    };
\addplot[
    color=blue
    ]
    coordinates {
    (100, 0.92010003)
    (101, 0.9159999)
    (102, 0.9163999)
    (103, 0.9157)
    (104, 0.9144999)
    (105, 0.9161999)
    (106, 0.9114)
    (107, 0.91660005)
    (108, 0.91620004)
    (109, 0.91409993)
    (110, 0.91520005)
    (111, 0.9124)
    (112, 0.9163)
    (113, 0.90939987)
    (114, 0.917)
    (115, 0.91440004)
    (116, 0.9178)
    (117, 0.91550004)
    (118, 0.91330004)
    (119, 0.91400003)
    (120, 0.9169999)
    (121, 0.9092001)
    (122, 0.9137001)
    (123, 0.9136999)
    (124, 0.91409993)
    (125, 0.9125999)
    (126, 0.9198)
    (127, 0.9211)
    (128, 0.92200005)
    (129, 0.9203)
    (130, 0.92050004)
    (131, 0.92080003)
    (132, 0.92149997)
    (133, 0.92030007)
    (134, 0.92190003)
    (135, 0.9218001)
    (136, 0.9213)
    (137, 0.9225)
    (138, 0.92200005)
    (139, 0.921)
    (140, 0.92050004)
    (141, 0.92200005)
    (142, 0.9217)
    (143, 0.9210999)
    (144, 0.9215)
    (145, 0.92140007)
    (146, 0.9219001)
    (147, 0.92159986)
    (148, 0.9201)
    (149, 0.92159986)
    (150, 0.9219999)
    (151, 0.9218)
    (152, 0.9224999)
    (153, 0.9218)
    (154, 0.9218)
    (155, 0.9225)
    (156, 0.9216)
    (157, 0.9224)
    (158, 0.92149997)
    (159, 0.9217)
    (160, 0.921)
    (161, 0.922)
    (162, 0.92149997)
    (163, 0.9228)
    (164, 0.92270005)
    (165, 0.92190003)
    (166, 0.92130005)
    (167, 0.9205001)
    (168, 0.9216)
    (169, 0.9204)
    (170, 0.92130005)
    (171, 0.9220999)
    (172, 0.92170006)
    (173, 0.9219)
    (174, 0.9225)
    (175, 0.92190003)

    };
\addplot[
    color=blue
    ]
    coordinates {
    (100, 0.91630006)
    (101, 0.91409993)
    (102, 0.91269994)
    (103, 0.91410005)
    (104, 0.9121001)
    (105, 0.91260004)
    (106, 0.9150999)
    (107, 0.91269994)
    (108, 0.9149)
    (109, 0.9125)
    (110, 0.9118999)
    (111, 0.911)
    (112, 0.91019994)
    (113, 0.9139)
    (114, 0.9149)
    (115, 0.9145)
    (116, 0.9148999)
    (117, 0.9121001)
    (118, 0.9129999)
    (119, 0.9107)
    (120, 0.91339993)
    (121, 0.91090006)
    (122, 0.9122001)
    (123, 0.91020006)
    (124, 0.91270006)
    (125, 0.9114001)
    (126, 0.9202001)
    (127, 0.91940004)
    (128, 0.9212001)
    (129, 0.921)
    (130, 0.9213)
    (131, 0.9222001)
    (132, 0.9222)
    (133, 0.9218001)
    (134, 0.92090005)
    (135, 0.9211)
    (136, 0.9218001)
    (137, 0.9215001)
    (138, 0.9222001)
    (139, 0.9208001)
    (140, 0.92230004)
    (141, 0.92090005)
    (142, 0.92140007)
    (143, 0.9216001)
    (144, 0.92230004)
    (145, 0.92230004)
    (146, 0.9206)
    (147, 0.9199)
    (148, 0.9214001)
    (149, 0.92240006)
    (150, 0.9228)
    (151, 0.92060006)
    (152, 0.9221)
    (153, 0.92230004)
    (154, 0.9239)
    (155, 0.9225)
    (156, 0.9234999)
    (157, 0.9214)
    (158, 0.9222)
    (159, 0.92139983)
    (160, 0.9219)
    (161, 0.9218)
    (162, 0.92140007)
    (163, 0.92190003)
    (164, 0.9223)
    (165, 0.9217999)
    (166, 0.92219996)
    (167, 0.9207)
    (168, 0.9226001)
    (169, 0.9225)
    (170, 0.9217)
    (171, 0.92210007)
    (172, 0.92310005)
    (173, 0.9221)
    (174, 0.9237)
    (175, 0.9235)
    };
    \addlegendentry{S-SGD}

\addplot[
    color=red
    ]
    coordinates {
    (100, 0.90139997)
    (101, 0.9089001)
    (102, 0.9082)
    (103, 0.90730006)
    (104, 0.90989995)
    (105, 0.90519994)
    (106, 0.9078001)
    (107, 0.9075999)
    (108, 0.8907)
    (109, 0.9057001)
    (110, 0.90459996)
    (111, 0.90190005)
    (112, 0.9008)
    (113, 0.9032)
    (114, 0.9036)
    (115, 0.91)
    (116, 0.9020999)
    (117, 0.90410006)
    (118, 0.9085001)
    (119, 0.9037999)
    (120, 0.90340006)
    (121, 0.90419996)
    (122, 0.90190005)
    (123, 0.90470004)
    (124, 0.90200007)
    (125, 0.9068001)
    (126, 0.919)
    (127, 0.9196)
    (128, 0.92020005)
    (129, 0.92079985)
    (130, 0.9202)
    (131, 0.9212999)
    (132, 0.9221)
    (133, 0.92270005)
    (134, 0.92140007)
    (135, 0.92149997)
    (136, 0.92219996)
    (137, 0.922)
    (138, 0.9221)
    (139, 0.92200005)
    (140, 0.9237)
    (141, 0.9221)
    (142, 0.92049986)
    (143, 0.9221)
    (144, 0.9225)
    (145, 0.92210007)
    (146, 0.92370003)
    (147, 0.922)
    (148, 0.9213)
    (149, 0.92300004)
    (150, 0.92140007)
    (151, 0.9210999)
    (152, 0.92219996)
    (153, 0.9211)
    (154, 0.9228)
    (155, 0.921)
    (156, 0.92010003)
    (157, 0.9202)
    (158, 0.9210999)
    (159, 0.9217)
    (160, 0.9214)
    (161, 0.92190003)
    (162, 0.9203)
    (163, 0.9200001)
    (164, 0.9203999)
    (165, 0.92090005)
    (166, 0.9216)
    (167, 0.9219)
    (168, 0.9219)
    (169, 0.9218)
    (170, 0.9210999)
    (171, 0.92080003)
    (172, 0.92240006)
    (173, 0.9200001)
    (174, 0.9208)
    (175, 0.9212)

    };
\addplot[
    color=red
    ]
    coordinates {
    (100, 0.90680003)
    (101, 0.90389997)
    (102, 0.90720004)
    (103, 0.90440005)
    (104, 0.9038999)
    (105, 0.90539986)
    (106, 0.9034999)
    (107, 0.90349996)
    (108, 0.9022999)
    (109, 0.90340006)
    (110, 0.90139997)
    (111, 0.90029997)
    (112, 0.90690005)
    (113, 0.9033001)
    (114, 0.9036001)
    (115, 0.90549994)
    (116, 0.90480006)
    (117, 0.90630007)
    (118, 0.90440005)
    (119, 0.9044001)
    (120, 0.90519994)
    (121, 0.90349996)
    (122, 0.8985999)
    (123, 0.9019001)
    (124, 0.8957999)
    (125, 0.9049001)
    (126, 0.91770005)
    (127, 0.921)
    (128, 0.92)
    (129, 0.91940004)
    (130, 0.9198)
    (131, 0.9214)
    (132, 0.9202)
    (133, 0.9196999)
    (134, 0.9197)
    (135, 0.92069983)
    (136, 0.9216)
    (137, 0.9196999)
    (138, 0.9203)
    (139, 0.9202999)
    (140, 0.9202)
    (141, 0.9193001)
    (142, 0.9213999)
    (143, 0.922)
    (144, 0.9204001)
    (145, 0.92159986)
    (146, 0.9204)
    (147, 0.92079985)
    (148, 0.9212)
    (149, 0.921)
    (150, 0.9194)
    (151, 0.9219)
    (152, 0.9213)
    (153, 0.9210999)
    (154, 0.92200005)
    (155, 0.92189986)
    (156, 0.9219)
    (157, 0.9220999)
    (158, 0.9223)
    (159, 0.92039984)
    (160, 0.92079985)
    (161, 0.9210999)
    (162, 0.92120004)
    (163, 0.92120004)
    (164, 0.9211)
    (165, 0.9210999)
    (166, 0.9216)
    (167, 0.92190003)
    (168, 0.9224)
    (169, 0.9213)
    (170, 0.92090005)
    (171, 0.9213)
    (172, 0.92300004)
    (173, 0.92120004)
    (174, 0.92149997)
    (175, 0.9216)

    };
\addplot[
    color=red
    ]
    coordinates {
    (100, 0.9023001)
    (101, 0.9036)
    (102, 0.90849996)
    (103, 0.9014999)
    (104, 0.9037001)
    (105, 0.90519994)
    (106, 0.9031001)
    (107, 0.9008)
    (108, 0.9024001)
    (109, 0.8988001)
    (110, 0.9012001)
    (111, 0.8922)
    (112, 0.8953)
    (113, 0.90230006)
    (114, 0.89909995)
    (115, 0.90190005)
    (116, 0.90200007)
    (117, 0.90009993)
    (118, 0.9017999)
    (119, 0.90269995)
    (120, 0.90160006)
    (121, 0.89910007)
    (122, 0.9006001)
    (123, 0.9018001)
    (124, 0.90450007)
    (125, 0.90549994)
    (126, 0.9178)
    (127, 0.9183001)
    (128, 0.9197)
    (129, 0.9206)
    (130, 0.91980004)
    (131, 0.9218)
    (132, 0.9208)
    (133, 0.92120004)
    (134, 0.91950005)
    (135, 0.9225001)
    (136, 0.9211)
    (137, 0.92090005)
    (138, 0.9221)
    (139, 0.9229)
    (140, 0.9228)
    (141, 0.92219996)
    (142, 0.92399997)
    (143, 0.921)
    (144, 0.9218)
    (145, 0.9230001)
    (146, 0.92270005)
    (147, 0.9215)
    (148, 0.92230004)
    (149, 0.92410004)
    (150, 0.9225001)
    (151, 0.9234)
    (152, 0.9215)
    (153, 0.92170006)
    (154, 0.92230004)
    (155, 0.92190003)
    (156, 0.92170006)
    (157, 0.92100006)
    (158, 0.9224)
    (159, 0.9223)
    (160, 0.92329985)
    (161, 0.9224)
    (162, 0.9213999)
    (163, 0.9214)
    (164, 0.9231)
    (165, 0.92120004)
    (166, 0.92200005)
    (167, 0.9215)
    (168, 0.9212999)
    (169, 0.9202999)
    (170, 0.92200005)
    (171, 0.921)
    (172, 0.92120004)
    (173, 0.92080003)
    (174, 0.91990006)
    (175, 0.922)

    };
\addplot[
    color=red
    ]
    coordinates {
    (100, 0.90669996)
    (101, 0.9082999)
    (102, 0.9082999)
    (103, 0.90419996)
    (104, 0.90480006)
    (105, 0.9067999)
    (106, 0.9103001)
    (107, 0.9056001)
    (108, 0.91040003)
    (109, 0.90440005)
    (110, 0.9048999)
    (111, 0.89979994)
    (112, 0.9046999)
    (113, 0.9056001)
    (114, 0.89659995)
    (115, 0.9029999)
    (116, 0.90489995)
    (117, 0.9018)
    (118, 0.89629996)
    (119, 0.9054999)
    (120, 0.9015)
    (121, 0.9025)
    (122, 0.9008999)
    (123, 0.90510005)
    (124, 0.9041001)
    (125, 0.9039001)
    (126, 0.916)
    (127, 0.9182999)
    (128, 0.9196999)
    (129, 0.91980004)
    (130, 0.91870004)
    (131, 0.91830003)
    (132, 0.9194)
    (133, 0.91960007)
    (134, 0.9196999)
    (135, 0.918)
    (136, 0.9202999)
    (137, 0.9212)
    (138, 0.9206999)
    (139, 0.9186)
    (140, 0.9198)
    (141, 0.9213999)
    (142, 0.9202)
    (143, 0.92050004)
    (144, 0.9212999)
    (145, 0.9204)
    (146, 0.9198)
    (147, 0.9190001)
    (148, 0.9201)
    (149, 0.9207)
    (150, 0.9216)
    (151, 0.9219)
    (152, 0.92090005)
    (153, 0.91980004)
    (154, 0.9215)
    (155, 0.92160004)
    (156, 0.92120004)
    (157, 0.92049986)
    (158, 0.92100006)
    (159, 0.92190003)
    (160, 0.9194001)
    (161, 0.92190003)
    (162, 0.9210999)
    (163, 0.9215001)
    (164, 0.9211)
    (165, 0.9220999)
    (166, 0.92200005)
    (167, 0.92160004)
    (168, 0.9219)
    (169, 0.9215)
    (170, 0.92120004)
    (171, 0.9217001)
    (172, 0.92050004)
    (173, 0.92)
    (174, 0.92010003)
    (175, 0.92190003)

    };
    \addlegendentry{SGD}

\end{axis}
\end{tikzpicture}
}
\caption{Convergence curve of the conventional SGD and S-SGD for ResNet20 on CIFAR-10. (a) Test accuracy plot from 0th to 90th epochs, (b) Test accuracy plot from 100th to 175th epochs.}
\label{fig:convergence_curve}
\end{figure}

\figurename~\ref{fig:convergence_curve} shows the convergence curves of SGD and S-SGD for ResNet20 on CIFAR-10 dataset. The training was conducted for 175 epochs. \figurename~\ref{fig:convergence_curve} (a) shows the beginning and intermediate stages of training, and \figurename~\ref{fig:convergence_curve} (b) depicts the final stage. Here, we can notice that both SGD and S-SGD convergences are similar at the beginning stage, but S-SGD outperforms SGD when applied to intermediate and final stages. 

\begin{table}[t]
\caption{Test accuracy (\%) for ResNet20 on CIFAR-10 and CIFAR-100 under the equal computation budget with SGD. S denotes that the symmetrical noises are injected.}
   \label{tab:compute_compare}
   \centering
\begin{tabular}{ccc}
\\\toprule
 Train Epochs & CIFAR-10 & CIFAR-100 \\ 
 \midrule
75 - 50 - 50 & 92.17 & 68.26 \\ 
37S - 25S - 25S & 91.70 & 67.89 \\ 
75 - 50 - 25S & 92.33 & 69.32 \\ 
75 -25S -50 & 92.06 & 68.99 \\ 
75 - 25S - 25S & 92.55 & 69.57 \\ \bottomrule
\end{tabular}
\end{table}

 In Table \ref{tab:compute_compare}, we show various mixes of SGD and S-SGD under the constraint that the computation budget does not increase compared to SGD training of 75 epochs, 50 epochs, and 50 epochs. For example, 75 epochs of SGD training at the beginning stage can be replaced by 37 epochs of S-SGD training because S-SGD consumes twice the computation of SGD for each epoch. Here, we can find that `75 - 25S - 25S' that means 75 epochs of SGD training for the beginning part, 25 epochs of S-SGD in the middle, and 25 epochs of S-SGD at the final stage of training shows the best results for both CIFAR-10 and CIFAR-100 datasets.

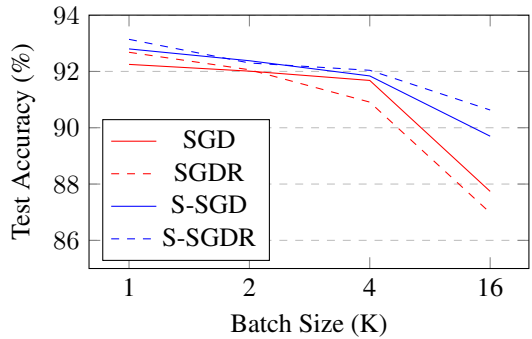
\begin{figure}[t]
\centering
\begin{tikzpicture}
\begin{axis}[
    xmode=log,
    log ticks with fixed point,
    log basis x={2},
    xlabel={Batch Size (K)},
    ylabel={Test Accuracy (\%)},
    xmin=0, xmax=10,
    ymin=85, ymax=94,
    height=0.6\linewidth,
    width=0.9\linewidth,
    xtick={1, 2, 4, 8},
    xticklabels={1, 2, 4, 16},
    ytick={86, 88, 90, 92, 94},
    legend pos=south west,
    ymajorgrids=true,
    grid style=dashed,
]
\addplot[
    color=red,
    mark=none,
    ]
    coordinates {
(1,92.25)
(2,92.01)
(4,91.68)
(8,87.74)
    };
    \addlegendentry{SGD}
\addplot[
    color=red,
    mark=none,
    dashed
    ]
    coordinates {
    (1,92.68)
(2,92.06)
(4,90.91)
(8,86.98)
    };
    \addlegendentry{SGDR}
    \addplot[
    color=blue,
    mark=none,
    ]
    coordinates {
    (1,92.80)
(2,92.38)
(4,91.84)
(8,89.70)
    };
    \addlegendentry{S-SGD}
        \addplot[
    color=blue,
    mark=none,
    dashed
    ]
    coordinates {
    (1,93.14)
(2,92.30)
(4,92.04)
(8,90.63)
    };
    \addlegendentry{S-SGDR}
    \end{axis}
\end{tikzpicture}
\caption{ Large-batch training of ResNet20 on CIFAR-10.}
\label{fig:large_batch}  
\end{figure}


\subsection{Large Batch Training}

Training a DNN using multi-GPUs is an attractive method for reducing the training time, and it typically requires a batch-size increase because a data-parallel approach is widely employed for reducing the communication overhead. The increased batch size decreases the data-dependent noise of the SGD method and hinders the escape from sharp minima. Recent studies suggest that increasing the learning rate proportionally to the batch size relieves this problem \cite{smith2017don}. However, the learning rate cannot be flexibly scaled owing to the stability problem \cite{you2017scaling}. 

We also evaluated the performance of the S-SGD method when combined with a learning rate scheduling method that is also intended to reach flat minima. Stochastic gradient descent with warm restarts (SGDR) was used \cite{loshchilov2016sgdr}. The SGDR periodically resets the learning rate to the initial value.
Specifically, we used an initial learning rate of 0.1 and decayed with the cosine function for the initial 10 epochs. The learning rate decay period was doubled at every new restart. The experimental results with ResNet20 on the CIFAR-10 dataset are shown in \figurename~\ref{fig:large_batch}. The S-SGDR method is the combination of the symmetrical weight noise injection and the SGDR learning rate scheduling. We used 300 epochs for the SGDR and S-SGDR methods.

Here, we conducted experiments to show that the proposed S-SGD or S-SGDR methods can improve the performance of large-batch training. We used a four-GPU training employing synchronous SGD using Horovod \cite{sergeev2018horovod}. The experiments were performed on NVIDIA DGX-1 \cite{dgx}. The batch size for each GPU was one quarter of that shown in \figurename~\ref{fig:large_batch}. Batch normalization statistics are computed for each device in training \cite{hoffer2017train}. The other hyperparameters were identical to those of the single-GPU training.

\figurename~\ref{fig:large_batch} shows the performance of ResNet20 on the CIFAR-10 dataset, in which batch sizes of 1, 2, 4, and 16 K were employed. For each batch size, four training methods, SGD, SGDR, S-SGD, and S-SGDR, were employed for performance measurement. For a batch size of 16,384, the proposed S-SGDR method showed the best test accuracy, approximately 2.9\% higher than that of the conventional SGD-based training. We found almost no performance decrease compared to that of the small-batch training even at a batch size of 4,096. The performance degradation with increasing batch size is very slow when S-SGDR or S-SGD is used. The training data size of CIFAR-10 dataset was only 50,000. Thus, the batch size of 4K was almost 1/10 of the total training dataset. 


\section{Experimental Results}\label{sec:exp}

We conducted various experiments to assess the performance of S-SGD algorithm with CNNs for image classification, Simple Gated ConvNet for speech recognition and RNNs for language modeling. All the experiments were implemented in TensorFlow \cite{abadi2016tensorflow}.

\subsection{Image Classification with CNN Models on CIFAR-10, CIFAR-100, and ImageNet}

We measured the performance of S-SGD on various CNN models using CIFAR-10, CIFAR-100, and ImageNet datasets \cite{russakovsky2015imagenet}. The S-SGD method employs noise strengths of 0.4 $\sigma$ and 0.5 $\sigma$. 

The experimental results are presented in Table \ref{tab:cifar_results}, for which the test accuracies of widely used CNN models were measured.  As shown in Table \ref{tab:cifar_results}, the S-SGD method exhibited approximately 0.45\% (92.62\% - 92.17\%) of accuracy increase for ResNet20 on CIFAR-10, which means that the error rate decreased from 7.83\% to 7.38\%. An accuracy increase of 0.44\% (93.50\% - 93.06\%) was also observed for ResNet56 on CIFAR-10. A higher performance gain, about 1.5\% to 2.0\%, can be seen for the CIFAR-100 dataset.   

\begin{table}[t]
\caption{ResNet \cite{he2016deep}, WideResNet \cite{zagoruyko2016wide} and VGG \cite{simonyan2014very} trained with CIFAR-10 and CIFAR-100 using conventional SGD and S-SGD. Average test accuracy values (\%) over 5 runs are reported. The value inside of the parentheses in training method indicates the noise level.}
   \label{tab:cifar_results}
   \centering
   \resizebox{\columnwidth}{!}{
\begin{tabular}{llcc}
\\\hline\hline
\textbf{Model} & \textbf{Method} & \textbf{CIFAR-10} & \textbf{CIFAR-100} \\ \hline
\multirow{3}{*}{ResNet20} & SGD & 92.17 & 68.00 \\ \cline{2-4} 
 & S-SGD (0.4) & 92.34 (+0.17) & 69.44 (+1.44)\\ \cline{2-4} 
 & S-SGD (0.5) & 92.62 (+0.45) & 69.48 (+1.48) \\ \hline\hline
\multirow{3}{*}{ResNet56} & SGD & 93.06 & 70.12 \\ \cline{2-4} 
 & S-SGD (0.4) & 93.50 (+0.44) & 71.61 (+1.49)\\ \cline{2-4} 
 & S-SGD (0.5) & 93.50 (+0.44) & 72.19 (+2.07)\\ \hline\hline
 \multirow{3}{*}{WRNet28-2} & SGD & 94.20 & 74.81 \\ \cline{2-4} 
 & S-SGD (0.5) & 94.50 (+0.30) & 75.43 (+0.62) \\ \cline{2-4} 
 & S-SGD (0.7) & 94.68 (+0.48) & 75.55 (+0.74)\\ \hline\hline
 \multirow{3}{*}{WRNet28-4} & SGD & 95.07 & 77.12 \\ \cline{2-4} 
 & S-SGD (0.5) & 95.20 (+0.13) & 77.66 (+0.54)\\ \cline{2-4} 
 & S-SGD (0.7) & 95.30 (+0.23) & 77.57 (+0.45)\\ \hline\hline
\multirow{3}{*}{VGG6} & SGD & 92.94 & 74.81 \\ \cline{2-4} 
 & S-SGD (0.7) & 94.30 (+1.36) & 75.61 (+0.80)\\ \cline{2-4} 
 & S-SGD (0.9) & 94.37 (+1.43) & 75.81 (+1.00)\\ \hline\hline
\multirow{3}{*}{VGG16} & SGD & 92.18 & 72.68 \\ \cline{2-4} 
 & S-SGD (0.7) & 93.72 (+1.54) & 73.31 (+0.63)\\ \cline{2-4} 
 & S-SGD (0.9) & 93.57 (+1.39) & 73.11 (+0.43)\\ \hline\hline
\end{tabular}
}
\end{table}

 \begin{table}[]
  \caption{Training of ResNet18 and ResNet50 on ImageNet. The value inside of the parentheses in training method indicates the noise level. The results are the average of three runs.}
  \label{tab:resnet18_imagenet}
  \centering
  \begin{tabular}{l l c c }
    \\\toprule
     \textbf{Model} & \textbf{Method} & \textbf{Top-1 Acc} & \textbf{Top-5 Acc}\\
    \midrule
      \multirow{2}{*}{ResNet18} &SGD &  70.24&  89.43 \\ 
      &S-SGD (0.25) & 70.42 & 89.57\\ 
    \midrule
      \multirow{2}{*}{ResNet50} &SGD &  76.11&  93.04 \\ 
      &S-SGD (0.25) & 76.38 & 93.13\\ 
      
    \bottomrule
  \end{tabular}  
\end{table}


The ResNet18 and ResNet50 training results on ImageNet dataset are exhibited in Table \ref{tab:resnet18_imagenet}. We used an 8-core Cloud TPU for training. A batch size of 1024, assigning 128 for each core, was used. Each model was trained for 90 epoch with learning rate scheduling from \cite{ying2018image}. 


\subsection{Speech Recognition with Simple Gated ConvNet}

\begin{table}[h]
  \caption{WER (\%) of Simple Gated ConvNet trained with S-SGD. The models are trained on WSJ si-284.} 
  \label{tab:sssgd_284}
  \centering
  \begin{tabular}{l c c }
    \\
    \toprule
    \multicolumn{1}{l}{\textbf{Model}}
    &\multicolumn{1}{c}{\textbf{Params.}}
                                         & \multicolumn{1}{c}{\textbf{WER (\%)}}\\
    \midrule
    12x190 SGCN   & 1.09M & 21.66\\
    12x190 SGCN  + S-SGD  & 1.09M & 19.90(-1.76)\\
    12x300 SGCN    & 2.24M & 18.30\\
    12x300 SGCN  + S-SGD  & 2.24M & 16.87(-1.43)\\ 
    \bottomrule
  \end{tabular}  
\end{table}
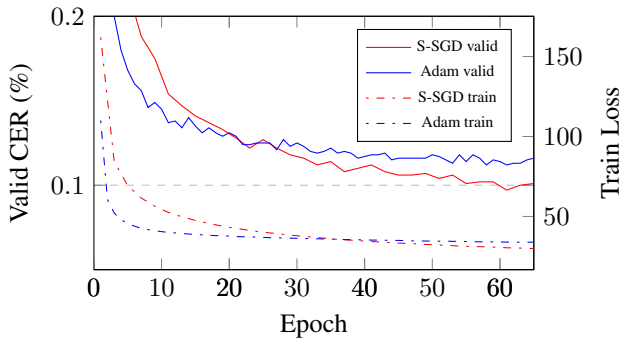
\begin{figure}[h]
\centering
\pgfplotsset{compat=1.10}
\begin{tikzpicture}
\begin{axis}[
    xlabel={Epoch},
    ylabel={Valid CER (\%)},
    xmin=0, xmax=65,
    ymin=0.05, ymax=0.2,
    height=0.6\linewidth,
    width=0.9\linewidth,
    xtick={0, 10, 20, 30, 40, 50, 60},
    ytick={0, 0.1, 0.2},
    legend style={at={(0.95,0.95)}},
    legend style={font=\fontsize{6}{7}\selectfont},
    ymajorgrids=true,
    grid style=dashed,
]
\addplot[
    color=red,
    mark=none,
    ] table {data/valid_cer_sssgd.txt};
    \label{plot_valid_S}
    \addlegendentry{S-SGD valid}
\addplot[
    color=blue,
    mark=none,
    ]table {data/valid_cer.txt};
    \label{plot_valid_A}
    \addlegendentry{Adam valid}
     \addlegendimage{/pgfplots/refstyle=train_loss_ssgd}\addlegendentry{S-SGD train}
\addlegendimage{/pgfplots/refstyle=valid_cer}\addlegendentry{Adam train}
\end{axis}

\begin{axis}[
    height=0.6\linewidth,
    width=0.9\linewidth,
    axis y line*=right,
    ylabel={Train Loss},
    xmin=0, xmax=65,
]
\addplot[
    dash pattern= on 1pt off 3pt on 3pt off 3pt,
    color=red
    ] table {data/train_loss_sssgd.txt};
    \label{train_loss_ssgd}
\addplot[
    dash pattern= on 1pt off 3pt on 3pt off 3pt,
    color=blue
    ] table {data/train_loss.txt};
    \label{valid_cer}
    
\end{axis}

\end{tikzpicture}
\caption{Train loss and valid CER curve of the SGCN. Solid and dashed lines denote the valid CER and train loss, respectively. }
\label{fig:fig_valid_curve}  
\end{figure}

The Simple Gated ConvNet (SGCN) is a convolution-based sequence modeling network used for end-to-end speech recognition \cite{lukas2019simple}. The connectionist temporal classification (CTC) loss is used for training \cite{graves2006connectionist}. Usually, end-to-end speech recognition using CTC loss employs long short-term memory (LSTM) \cite{hochreiter1997long} based models, but SGCN is more suitable for embedded applications because it is free from the sequential dependency problem inherent to LSTM RNN models, and allows the processing of multiple output samples at a time. The network configuration and model parameters were imported from \cite{lukas2019simple}.The SGCN model consists of 12 layers and each layer contains 190 units. Consequently, the number of parameters is about 1 M. The SGCN employs 1-dimensional time-depth convolution to increase the sequential classification capability without considerably increasing the number of parameters. Wall Street Journal si-284 \cite{paul1992design} is used for the training, which contains 81 hours of training data. The original work was trained using Adam. We trained the network using S-SGD Adam, and compared the results. 

\figurename~\ref{fig:fig_valid_curve} shows the training loss and validation character error rate (CER) curves when the model is trained with Adam and S-SGD Adam. The training results are summarized in Table \ref{tab:sssgd_284}. For the 1M parameter SGCN model, the word error rate (WER) of Adam training with greedy decoding was 21.66\%, while the S-SGD result was 19.90\%. A similar performance improvement was observed for the 2M parameter model. This clearly indicates that S-SGD is very effective for sequence recognition problems.

\subsection{Language Modeling with LSTM RNN}

We trained an LSTM network on the Penn Tree Bank (PTB) dataset \cite{marcus1994penn} for word-level language modeling. This dataset contains about one million words divided into training, validation, and test sets of about 930K, 74K and 82K words, respectively, with a vocabulary size of 10K. 
The network consists of two layers, each with 1,500 hidden units, resulting in about 6.7 million weights. We unrolled the network in 35 time steps for training. Dropout was applied between the layers. We reproduced the training pipeline of \cite{zaremba2014recurrent} for this network (SGD without momentum) and obtained a word perplexity of around 81.43 and 78.68 on the validation and test sets, respectively, with this setup; these numbers closely match the results of the original work.
We ran S-SGD for 55 epochs, and obtained a word perplexity of 76.86 and 73.83 on the validation and test sets, respectively. The performance according to the injected noise level is shown in Table \ref{tab:RNN_LM}. We observed that S-SGD works very well with RNN.
Table \ref{tab:RNN_LM} also shows the results of the small RNN model, which consists of a single LSTM layer containing 300 units \cite{hubara2017quantized}. In both models, we could obtain very improved results when compared to original RNN-based language models.It is well-known that the Dropout technique does not yield good results when applied to recurrent paths of RNN, and in the original work of \cite{zaremba2014recurrent} the dropout was only applied to the forward paths. In our implementation, we injected noises to all weights, except for the biases. Thus, in the forward-path, both Dropout and symmetrical weight noise injection were applied, while in the recurrent path, only symmetrical weight noise injection was used.

 \begin{table}[th]
  \caption{Test perplexity of RNN based language modeling on PTB dataset. The value inside of the parentheses in training method indicates the noise level. }
  \label{tab:RNN_LM}
  \centering
  \begin{tabular}{l l c c }
    \\\toprule
     \textbf{Model} & \textbf{Method} & \textbf{Test Perplexity} \\
    \midrule
      \multirow{4}{*}{Large RNN} &SGD &  78.68\\ 
      &S-SGD (0.8) & 74.47 (-4.21) \\ 
      &S-SGD (0.9) & 73.95 (-4.73)\\
      &S-SGD (1.0) & 73.83 (-4.85)\\
    \midrule
      \multirow{4}{*}{Small RNN} &SGD &  88.85\\ 
      &S-SGD (0.6) & 85.56 (-3.29)\\  
      &S-SGD (0.7) & 84.86 (-3.99)\\    
      &S-SGD (0.8) & 86.04 (-2.81)\\   
    \bottomrule
  \end{tabular}  
\end{table}

\begin{figure*}[t]
    \centering
    \begin{minipage}[t]{0.45\linewidth}
      \begin{center}
        \includegraphics[width=0.8\linewidth]{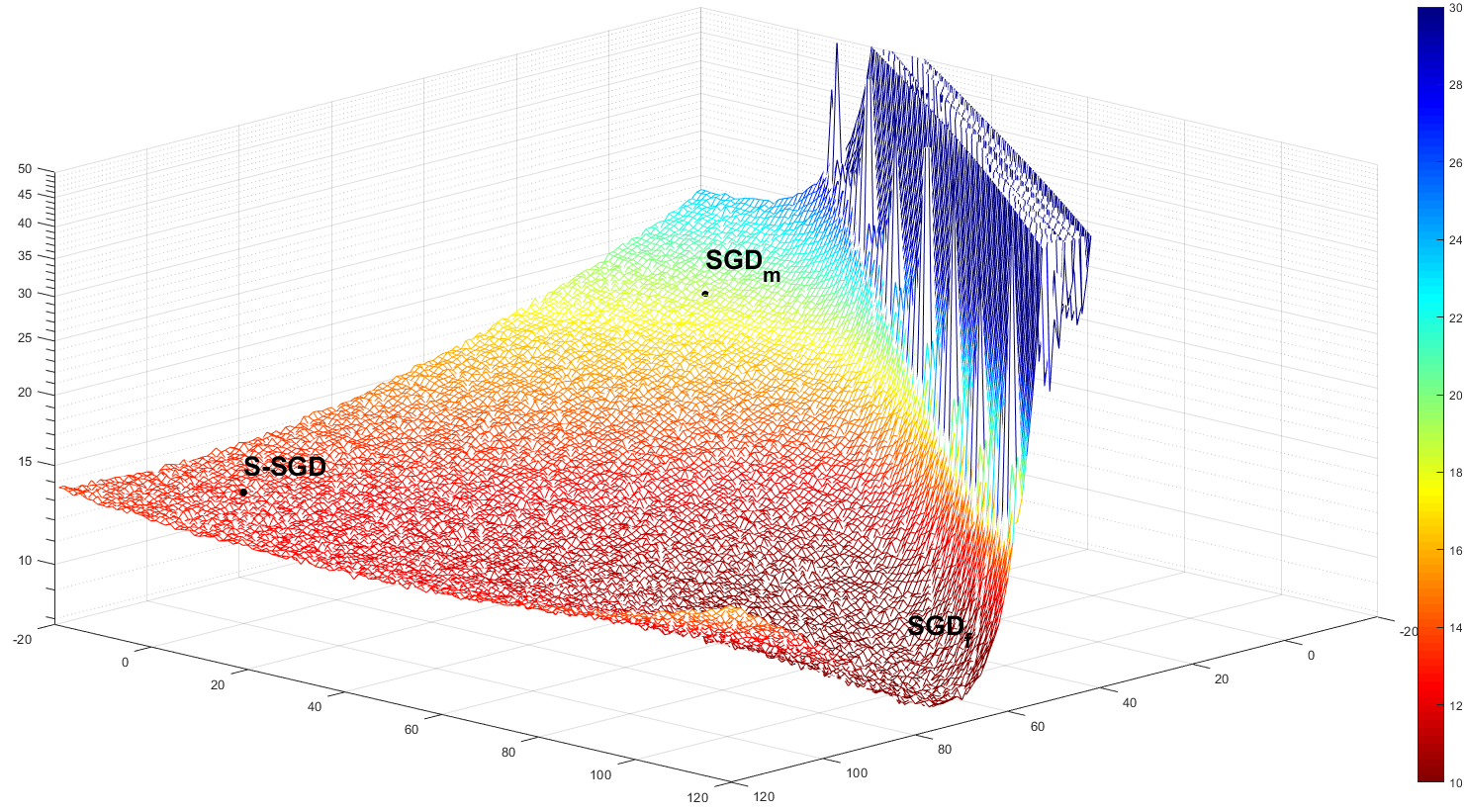} \\
            (a) Train
      \end{center}
    \end{minipage}
    \begin{minipage}[t]{0.45\linewidth}
      \begin{center}
        \includegraphics[width=0.8\columnwidth]{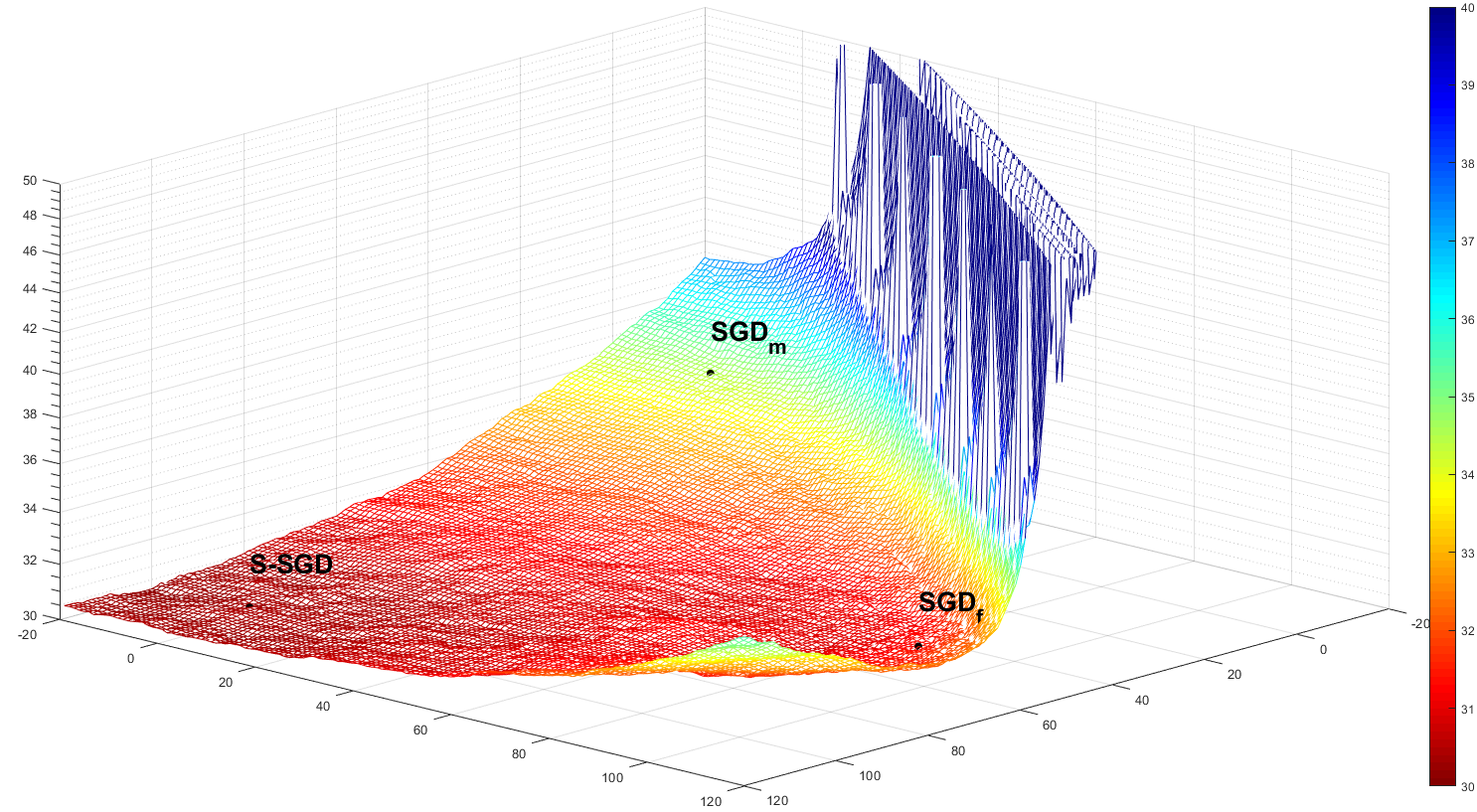} \\
        (b) Test
      \end{center}
    \end{minipage}
    
    \caption{Illustrations of loss surface for train and test error on ResNet20 CIFAR-100. The three points are ResNet20 that trained with 125 epochs of SGD ($\text{SGD}_\text{m}$), $\text{SGD}_\text{m}$ + 50 epochs of SGD ($\text{SGD}_\text{f}$), and $\text{SGD}_\text{m}$ + 50 epochs of S-SGD ($\text{S-SGD}$).}
    \label{fig_graphic_3dloss}
\end{figure*}

\section{Discussion}\label{sec:discussion}

\subsection{Comparison with other regularization methods}
Most DNNs currently used employ a large number of parameters, but the amount of training data is insufficient for a good training. Thus, the importance of regularization grows as the network size increases.  We compared the proposed S-SGD with the well-known regularization methods, such as Dropout, DropConnect, and SmoothOut. 

Dropout is an approach to regularization in neural networks that helps to reduce interdependent learning among the neurons \cite{srivastava2014dropout}. Individual nodes are either dropped out of the net or kept, with probabilities $1-p$ and $p$, respectively. Incoming and outgoing edges to a dropped-out node are also removed. Dropout is especially useful when applied to fully connected layers, but it has limitations when applied to batch normalization layers in CNNs or recurrent paths in RNNs \cite{ioffe2015batch, zaremba2014recurrent}. In Section \ref{sec:optim_noise}, we showed the performance improvement of S-SGD training for ResNet20 using CIFAR-100 dataset. However, for the same setup, Dropout applied to the fully connected layer did not yield any performance improvements regardless of the dropout rate. 

DropConnect sets a randomly selected subset of weights within the network to zero \cite{pmlr-v28-wan13}. Thus, it can be considered a fine-grained version of the Dropout. DropConnect was originally developed for the regularization of fully connected layers. RNN regularization with DropConnect was studied in \cite{merity2018regularizing}. We applied DropConnect to the ResNet20 model using CIFAR-10 data. When DropConnect was applied to convolution layers, the training was not successful, resulting in very low accuracy. Application of DropConnect to the fully connected layers resulted in 92.18\% of accuracy, which was, however, just comparable to SGD (92.17\%) and lower than  S-SGD(92.62\%).

The SmoothOut is a weight noise injection method in which the noise injected weights are used for forward- and backward- passes \cite{wen2018smoothout}. As the noise is randomly distributed, the operation may be considered an analog version of the DropConnect. The comparison of S-SGD with SmoothOut was presented in Section \ref{sec:optimize}. There are no application results of SmoothOut on RNNs. 


\subsection{Visualization of Loss Surface} 
\label{secsec_visualization}

We attempted to observe the training and test loss surfaces of SGD and S-SGD trained models, using the visualization method developed by~\cite{garipov2018loss}. This method can show the locations of three different models. We trained ResNet20 on CIFAR-100 dataset using SGD for the first 75 and the next 50 epochs with learning rates (LR) of 0.1 and 0.01, respectively. This half-trained model was stored as $\text{SGD}_\text{m}$. The $\text{SGD}_\text{m}$ was trained using SGD with LR of 0.001 for 50 epochs to obtain the fully trained ResNet20. This model was stored as $\text{SGD}_\text{f}$. In addition, $\text{SGD}_\text{m}$ was trained using S-SGD with LR of 0.001 for 50 epochs, and it was named $\text{S-SGD}$. The training accuracy values of $\text{SGD}_\text{m}$, $\text{SGD}_\text{f}$, and $\text{S-SGD}$ were 81.44\%, 91.14\%, and 86.62\%, respectively. The test accuracy values of $\text{SGD}_\text{m}$, $\text{SGD}_\text{f}$, and $\text{S-SGD}$ were 65.55\%, 68.51\%, and 69.63\%. 

\figurename~\ref{fig_graphic_3dloss} (a) shows the training loss surface, where $\text{SGD}_\text{m}$, $\text{SGD}_\text{f}$, and $\text{S-SGD}$ are located. We observed that $\text{SGD}_\text{f}$ is near the lowest point in the loss surface. This seems natural considering the operation of SGD seeking minimum. However, we found that $\text{SGD}_\text{f}$ is located very close to the very steep loss wall, suggesting the possibility of sharp minimum. The loss would sharply increase with variation of input data characteristics, which means overfitting or poor generalization. For training set, loss of $\text{S-SGD}$ is a little higher than that of $\text{SGD}_\text{f}$ but it is located far from the steep loss wall, which shows the possibility of improved robustness to input data perturbation. 
\figurename~\ref{fig_graphic_3dloss} (b) exhibits the test loss surface, where $\text{SGD}_\text{m}$, $\text{SGD}_\text{f}$ and $\text{S-SGD}$ are included. The location of $\text{S-SGD}$ becomes the minimum point located near the center of the basin. We noticed that S-SGD avoids overfitting by adding weight noise in the training phase. The amount of added noise would determine the distance to the sharp loss wall. \figurename~\ref{fig:fig_ssgd} is a 1-D sectional view of \figurename~\ref{fig_graphic_3dloss}, where the line connecting $\text{SGD}_\text{f}$ and $\text{S-SGD}$ becomes the axis of 1-D plot. 

We have included all the training parameters and training results of the experiments in the Appendix. The eigenvalues of Hessian for SGD and S-SGD trained networks are also presented in the Appendix. 




\section{Concluding Remarks}\label{sec:conclude}

We proposed the Symmetrical Stochastic Gradient Descent (S-SGD) algorithm, which injects symmetrical weight noises for measuring loss surface flatness in DNN training. This method aids in reaching flat minima during SGD training because the loss values are measured at two points at each weight update. The S-SGD method is very stable because two symmetrical noises are added to the weights. 
The weight update analysis for this algorithm shows that the gradient for the weight update depends on the flatness of loss surface, or Hessian. We conducted many experiments using several CNN models, datasets, batch sizes, and learning rate scheduling for a performance evaluation of the method. The S-SGD also showed substantial performance improvements when applied to end-to-end speech recognition with Simple Gated ConvNet and language modeling with LSTM RNN. When compared to previously developed regularization methods, such as Dropout and DropConnect, S-SGD is considerably more flexibile and can be applied to fully connected, convolution and recurrent layers.

\subsection*{Acknowledgment}

This work was supported in part by AIR Lab (AI Research Lab) in Hyundai Motor Company
through HMC-SNU AI Consortium Fund. This work was also supported in part 
by the National Research Foundation of Korea (NRF) grant funded by the Korea
government (MSIP) (No. 2018R1A2A1A05079504). 





\bibliography{main}
\bibliographystyle{icml2020}



\end{document}


\icmltitlerunning{}

\icmltitle{S-SGD: Symmetrical Stochastic Gradient Descent\\for Reaching Flat Minima in Deep Neural Network Training \\ -Appendix-}

\appendix
\section{Additional Results on CIFAR-10}
Table 2 of the main text shows the training summary of various CNN models on CIFAR-10 dataset. Here, we present the detailed results on five or three runs.


\begin{table}[h]
\caption{Test accuracy on CIFAR-10 using conventional SGD and S-SGD. Please refer to Table 2 in the main paper. The value inside of the parentheses in training method indicates
the noise strength. }
\vskip 0.05in
\centering \resizebox{0.9\textwidth}{!}{
\begin{tabular}{|l|l|l|l|l|l|l|l|l|l|}
\hline
Model                           & Method      & 1st    & 2nd    & 3rd    & 4th    & 5th    & Average     & Min    & Max    \\ \hline
\multirow{3}{*}{ResNet20}       & SGD         & 0.9195 & 0.9226 & 0.923  & 0.921  & 0.9206 & 0.9213     & 0.9195 & 0.923  \\ \cline{2-10} 
                                & S-SGD (0.5) & 0.9292 & 0.9258 & 0.9237 & 0.9259 & 0.9242 & 0.9258     & 0.9237 & 0.9292 \\ \cline{2-10} 
                                & S-SGD (0.4) & 0.9222 & 0.9235 & 0.9245 & 0.9262 & 0.9222 & 0.9237     & 0.9222 & 0.9262 \\ \hline
\multirow{3}{*}{ResNet56}       & SGD         & 0.9277 & 0.9326 & 0.9324 & 0.9265 & 0.9339 & 0.9306     & 0.9265 & 0.9339 \\ \cline{2-10} 
                                & S-SGD (0.5) & 0.9354 & 0.9247 & 0.9344 & 0.9406 & 0.94   & 0.9350     & 0.9247 & 0.9406 \\ \cline{2-10} 
                                & S-SGD (0.4) & 0.9393 & 0.935  & 0.9343 & 0.9331 & 0.9332 & 0.9350     & 0.9331 & 0.9393 \\ \hline
\multirow{3}{*}{VGG6}           & SGD         & 0.9259 & 0.9268 & 0.9349 & 0.9275 & 0.9317 & 0.9294     & 0.9259 & 0.9349 \\ \cline{2-10} 
                                & S-SGD (0.7) & 0.9425 & 0.9421 & 0.9421 & 0.9467 & 0.9415 & 0.9430     & 0.9415 & 0.9467 \\ \cline{2-10} 
                                & S-SGD (0.9) & 0.9459 & 0.9451 & 0.9427 & 0.9404 & 0.9442 & 0.9437     & 0.9404 & 0.9459 \\ \hline
\multirow{3}{*}{VGG16}          & SGD         & 0.9244 & 0.9215 & 0.9191 & 0.9215 & 0.9227 & 0.9218     & 0.9191 & 0.9244 \\ \cline{2-10} 
                                & S-SGD (0.7) & 0.9391 & 0.9378 & 0.9358 & 0.9354 & 0.9377 & 0.9372     & 0.9354 & 0.9391 \\ \cline{2-10} 
                                & S-SGD (0.9) & 0.9345 & 0.9359 & 0.9345 & 0.9388 & 0.935  & 0.9357     & 0.9345 & 0.9388 \\ \hline
\multirow{3}{*}{WideResNet28-2} & SGD         & 0.9418 & 0.9435 & 0.9408 &        &        & 0.9420 & 0.9408 & 0.9435 \\ \cline{2-10} 
                                & S-SGD (0.5) & 0.9436 & 0.946  & 0.9454 &        &        & 0.945       & 0.9436 & 0.946  \\ \cline{2-10} 
                                & S-SGD (0.7) & 0.9473 & 0.9434 & 0.9498 &        &        & 0.9468 & 0.9434 & 0.9498 \\ \hline
\multirow{3}{*}{WideResNet28-4} & SGD         & 0.9504 & 0.9513 & 0.9505 &        &        & 0.9507 & 0.9504 & 0.9513 \\ \cline{2-10} 
                                & S-SGD (0.5) & 0.9538 & 0.9505 & 0.9517 &        &        & 0.952       & 0.9505 & 0.9538 \\ \cline{2-10} 
                                & S-SGD (0.7) & 0.9543 & 0.9529 & 0.9519 &        &        & 0.9530 & 0.9519 & 0.9543 \\ \hline
\end{tabular} }
\end{table}

\section{Detailed Results on CIFAR-100}
We conducted experiments on CIFAR-100 dataset. Training setup is identical to experiments on CIFAR-10 dataset as described in Section A.

\begin{table}[H]
\caption{Test accuracy on CIFAR-100 using conventional SGD and S-SGD. Please refer to Table 2 in the main paper. The value inside of the parentheses in training method indicates
the noise strength.}
\vskip 0.05in\centering \resizebox{0.9\textwidth}{!}{
\begin{tabular}{|l|l|l|l|l|l|l|l|l|l|}
\hline
Model                           & Method      & 1st    & 2nd    & 3rd    & 4th    & 5th    & Average     & Min    & Max    \\ \hline
\multirow{3}{*}{ResNet20}       & SGD         & 0.6843 & 0.6772 & 0.684  & 0.6759 & 0.6787 & 0.6800     & 0.6759 & 0.6843 \\ \cline{2-10} 
                                & S-SGD (0.5) & 0.6947 & 0.6881 & 0.6936 & 0.7008 & 0.6969 & 0.6948     & 0.6881 & 0.7008 \\ \cline{2-10} 
                                & S-SGD (0.4) & 0.6995 & 0.6936 & 0.6893 & 0.6914 & 0.6982 & 0.6944      & 0.6893 & 0.6995 \\ \hline
\multirow{3}{*}{ResNet56}       & SGD         & 0.7069 & 0.6997 & 0.7094 & 0.6958 & 0.694  & 0.7012     & 0.694  & 0.7094 \\ \cline{2-10} 
                                & S-SGD (0.5) & 0.7246 & 0.7261 & 0.7193 & 0.7255 & 0.7138 & 0.7219     & 0.7138 & 0.7261 \\ \cline{2-10} 
                                & S-SGD (0.4) & 0.7212 & 0.7089 & 0.7262 & 0.7227 & 0.7014 & 0.7161     & 0.7014 & 0.7262 \\ \hline
\multirow{3}{*}{VGG6}           & SGD         & 0.747  & 0.7476 & 0.7484 & 0.7476 & 0.7498 & 0.7481     & 0.747  & 0.7498 \\ \cline{2-10} 
                                & S-SGD (0.7) & 0.76   & 0.754  & 0.7532 & 0.7561 & 0.7572 & 0.7561      & 0.7532 & 0.76   \\ \cline{2-10} 
                                & S-SGD (0.9) & 0.7561 & 0.7609 & 0.7566 & 0.7607 & 0.756  & 0.7581     & 0.756  & 0.7609 \\ \hline
\multirow{3}{*}{VGG16}          & SGD         & 0.7284 & 0.7268 & 0.7264 & 0.7267 & 0.7255 & 0.7268     & 0.7255 & 0.7284 \\ \cline{2-10} 
                                & S-SGD (0.7) & 0.7321 & 0.7305 & 0.7356 & 0.7337 & 0.7337 & 0.7331     & 0.7305 & 0.7356 \\ \cline{2-10} 
                                & S-SGD (0.9) & 0.7325 & 0.7307 & 0.7309 & 0.7314 & 0.7301 & 0.7311     & 0.7301 & 0.7325 \\ \hline
\multirow{3}{*}{WideResNet28-2} & SGD         & 0.7481 & 0.7485 & 0.7392 &        &        & 0.7452 & 0.7392 & 0.7485 \\ \cline{2-10} 
                                & S-SGD (0.5) & 0.7543 & 0.7516 & 0.7531 &        &        & 0.753       & 0.7516 & 0.7543 \\ \cline{2-10} 
                                & S-SGD (0.7) & 0.7555 & 0.757  & 0.755  &        &        & 0.7558 & 0.755  & 0.757  \\ \hline
\multirow{3}{*}{WideResNet28-4} & SGD         & 0.7712 & 0.7721 & 0.7722 &        &        & 0.7718 & 0.7712 & 0.7722 \\ \cline{2-10} 
                                & S-SGD (0.5) & 0.7766 & 0.7768 & 0.779  &        &        & 0.7774 & 0.7766 & 0.779  \\ \cline{2-10} 
                                & S-SGD (0.7) & 0.7757 & 0.7754 & 0.7801 &        &        & 0.7770 & 0.7754 & 0.7801 \\ \hline
\end{tabular} }
\end{table}

\section{Additional Results on CIFAR-100 According to The Noise Level and Training Methods} 
We report experiments on CIFAR-100 dataset according to different noise levels and training methods. S-SGD Conv only and S-SGD Dense only denote the results of the selective noise injection. Dropout is only applied to the dense layer in SGD+Dropout. SmoothOut is explained in Section 4.1 in the main paper. S-SGD x2 and SmoothOut x2 are the results that employ two different noises for more precise estimation. These schemes demand twice the operations of the original methods, but these do not show improvements.  
\begin{figure*}[h]
\centering
\begin{tikzpicture}
\begin{axis}[
    xlabel={Noise level [$\sigma$]},
    ylabel={Accuracy[\%]},
    xmin=0, xmax=1,
    ymin=62, ymax=70,
    height=0.65\linewidth,
    width=1.0\linewidth,
    xtick={0,0.2,0.4,0.6,0.8,1.0},
    ytick={62,63,64,65,66,67,68,69,70},
    legend pos=south west,
    legend style={font=\fontsize{8}{9}\selectfont},
    ymajorgrids=true,
    grid style=dashed,
]
\addplot[color=red, mark=square,] coordinates {
(0   ,67.99)(	0.1	,	68.334	)(	0.2	,	68.624	)
(	0.3	,	69.196	)(	0.4	,	69.416	)(	0.5	,	69.398	)
(	0.6	,	68.792	)(	0.7	,	68.296	)(	0.8	,	67.108	)
(	0.9	,	65.4	)(	1	,	64.526	)
};\addlegendentry{S-SGD}
\addplot[color=blue, mark=square, 
] coordinates {
(0   ,67.99)(	0.1	,	68.425	)(	0.2	,	68.725	)
(	0.3	,	69.12	)(	0.4	,	69.32	)(	0.5	,	69.28	)
(	0.6	,	69.1	)(	0.7	,	68.3	)(	0.8	,	67.72	)
(	0.9	,	66.68	)(	1	,	65.36	)
};\addlegendentry{S-SGD Conv only}
\addplot[color=green, mark=square, 
]coordinates {
(0   ,67.99)(	0.1	,	68.18	)(	0.2	,	68.16	)
(	0.3	,	68.22	)(	0.4	,	68.18	)(	0.5	,	67.88	)
(	0.6	,	67.8	)(	0.7	,	68.18	)(	0.8	,	68.12	)
(	0.9	,	67.68	)(	1	,	67.72	)
};\addlegendentry{S-SGD Dense only}
\addplot[color=Sepia, mark=square,] coordinates {
(0   ,67.99)(	0.1	,	68.118	)(	0.2	,	68.308	)
(	0.3	,	68.694	)(	0.4	,	68.72	)(	0.5	,	68.264	)
(	0.6	,	67.738	)(	0.7	,	66.662	)(	0.8	,	65.92	)
(	0.9	,	64.79	)(	1	,	62.878	)
};\addlegendentry{SmoothOut}
\addplot[color=Orange, mark=square,] coordinates {
(0   ,67.99)(	0.1	,	68.22	)(	0.2	,	68.86	)
(	0.3	,	68.68	)(	0.4	,	69.18	)(	0.5	,	68.86	)
(	0.6	,	68.84	)(	0.7	,	67.96	)(	0.8	,	66.68	)
(	0.9	,	65.82	)(	1	,	64.16	)
};
    \addlegendentry{S-SGD x2}
\addplot[color=yellow, mark=square,] coordinates {
(0   ,67.99)(	0.1	,	68.46	)(	0.2	,	68.48	)
(	0.3	,	68.9	)(	0.4	,	68.98	)(	0.5	,	68.26	)
(	0.6	,	68.06	)(	0.7	,	66.56	)(	0.8	,	65.7	)
(	0.9	,	63.8	)(	1	,	62.64	)
};\addlegendentry{SmoothOut x2}
\addplot[dashed]coordinates{(0,67.4)(1,67.4)};\addlegendentry{SGD+Dropout(0.2)}
\addplot[name path=convOnly_p, draw=none] coordinates{
(0  ,   68.38) (	0.1	,	68.77571356	)(	0.2	,	69.12873258	)
(	0.3	,	69.38832816	)(	0.4	,	69.66205263	)(	0.5	,	69.42832397	)
(	0.6	,	69.22247449	)(	0.7	,	68.62403703	)(	0.8	,	67.97884358	)
(	0.9	,	67.05013511	)(	1	,	65.63018512	)

};
\addplot[name path=convOnly_m, draw=none] coordinates{
(0  ,   67.6) (	0.1	,	68.07428644	)(	0.2	,	68.32126742	)
(	0.3	,	68.85167184	)(	0.4	,	68.97794737	)(	0.5	,	69.13167603	)
(	0.6	,	68.97752551	)(	0.7	,	67.97596297	)(	0.8	,	67.46115642	)
(	0.9	,	66.30986489	)(	1	,	65.08981488	)

};
\addplot[color=blue, opacity=0.2] fill between[of=convOnly_p and convOnly_m];
\addplot[name path=denseOnly_p, draw=none] coordinates{
(0  ,   68.38)(	0.1	,	68.53637059	)(	0.2	,	68.56373258	)
(	0.3	,	68.70166378	)(	0.4	,	68.35888544	)(	0.5	,	68.19937439	)
(	0.6	,	68.18729833	)(	0.7	,	68.46635642	)(	0.8	,	68.51623226	)
(	0.9	,	68.02205263	)(	1	,	68.29183914	)
};
\addplot[name path=denseOnly_m, draw=none] coordinates{
(0  ,   67.6)(	0.1	,	67.82362941	)(	0.2	,	67.75626742	)
(	0.3	,	67.73833622	)(	0.4	,	68.00111456	)(	0.5	,	67.56062561	)
(	0.6	,	67.41270167	)(	0.7	,	67.89364358	)(	0.8	,	67.72376774	)
(	0.9	,	67.33794737	)(	1	,	67.14816086	)
};
\addplot[color=green, opacity=0.2] fill between[of=denseOnly_p and denseOnly_m];
\addplot[name path=ssgd_p, draw=none] coordinates{
(0  ,   68.38)(	0.1	,	68.53655863	)(	0.2	,	68.88919804	)
(	0.3	,	69.51100794	)(	0.4	,	69.66515858	)(	0.5	,	69.63542367	)
(	0.6	,	69.09211664	)(	0.7	,	68.66018402	)(	0.8	,	67.60086915	)
(	0.9	,	66.11940948	)(	1	,	65.06542562	)
};
\addplot[name path=ssgd_n, draw=none] coordinates{
(0  ,   67.6)(	0.1	,	68.13144137	)(	0.2	,	68.35880196	)
(	0.3	,	68.88099206	)(	0.4	,	69.16684142	)(	0.5	,	69.16057633	)
(	0.6	,	68.49188336	)(	0.7	,	67.93181598	)(	0.8	,	66.61513085	)
(	0.9	,	64.68059052	)(	1	,	63.98657438	)
};
\addplot[color=red, opacity=0.2] fill between[of=ssgd_p and ssgd_n];
\addplot[name path=smooth_p, draw=none] coordinates{
(0  ,   68.38)(	0.1	,	68.32310973	)(	0.2	,	68.66257016	)
(	0.3	,	69.17115825	)(	0.4	,	69.12981703	)(	0.5	,	68.58885381	)
(	0.6	,	68.16411031	)(	0.7	,	67.18182689	)(	0.8	,	66.42857644	)
(	0.9	,	65.20731283	)(	1	,	63.61009972	)
};
\addplot[name path=smooth_n, draw=none] coordinates{
(0  ,   67.6)(	0.1	,	67.91289027	)(	0.2	,	67.95342984	)
(	0.3	,	68.21684175	)(	0.4	,	68.31018297	)(	0.5	,	67.93914619	)
(	0.6	,	67.31188969	)(	0.7	,	66.14217311	)(	0.8	,	65.41142356	)
(	0.9	,	64.37268717	)(	1	,	62.14590028	)
};
\addplot[color=Sepia, opacity=0.2] fill between[of=smooth_p and smooth_n ];
\addplot[name path=ssgdx2_p, draw=none] coordinates{
(0  ,   68.38)(	0.1	,	68.39888544	)(	0.2	,	69.16495901	)
(	0.3	,	69.28580525	)(	0.4	,	69.53637059	)(	0.5	,	69.18093613	)
(	0.6	,	69.18351128	)(	0.7	,	68.17908902	)(	0.8	,	66.763666	)
(	0.9	,	66.29644517	)(	1	,	64.73271284	)
};
\addplot[name path=ssgdx2_m, draw=none] coordinates{
(0  ,   67.6)(	0.1	,	68.04111456	)(	0.2	,	68.55504099	)
(	0.3	,	68.07419475	)(	0.4	,	68.82362941	)(	0.5	,	68.53906387	)
(	0.6	,	68.49648872	)(	0.7	,	67.74091098	)(	0.8	,	66.596334	)
(	0.9	,	65.34355483	)(	1	,	63.58728716	)
};
\addplot[color=Orange, opacity=0.2] fill between[of=ssgdx2_p and ssgdx2_m ];
\addplot[name path=smoothx2_p, draw=none] coordinates{
(0  ,   68.38)(	0.1	,	68.84470768	)(	0.2	,	68.92384682	)
(	0.3	,	68.97071068	)(	0.4	,	69.32205263	)(	0.5	,	68.54809721	)
(	0.6	,	68.25493589	)(	0.7	,	66.80083189	)(	0.8	,	65.93452079	)
(	0.9	,	64.28989795	)(	1	,	63.38699398	)
};
\addplot[name path=smoothx2_m, draw=none] coordinates{
(0  ,   67.6)(	0.1	,	68.07529232	)(	0.2	,	68.03615318	)
(	0.3	,	68.82928932	)(	0.4	,	68.63794737	)(	0.5	,	67.97190279	)
(	0.6	,	67.86506411	)(	0.7	,	66.31916811	)(	0.8	,	65.46547921	)
(	0.9	,	63.31010205	)(	1	,	61.89300602	)
};
\addplot[color=yellow, opacity=0.2] fill between[of=smoothx2_p and smoothx2_m ];
    \end{axis}
\end{tikzpicture}
\caption{Test accuracy according to the amount of injected noise in S-SGD and SmoothOut. Average test accuracy value and its error bar are computed from 5 runs. Please refer to Figure 3 in the main paper.}
\label{fig:fig_bell_curve}  
\end{figure*}
\newpage
\section{Additional Results on PTB Language Model}
Table 5 of the main paper shows the performance of two LSTM RNN based language models, trained using the Penn Tree Bank (PTB) dataset. Here we show the test perplexity (PPL) when the noise level varies. The detailed training setup can be found at Section 5.3 of the main-text. The large network consists of two layers, each with 1,500 hidden units, resulting in about 6.7 million weights. The small RNN network consists of a single LSTM layer containing 300 units.


\begin{figure*}[h] \centering
\begin{tikzpicture}
\begin{axis}[
    xlabel={Noise level [$\sigma$]},
    ylabel={Test PPL},
    xmin=0, xmax=1.4,
    ymin=65, ymax=90,
    height=0.55\linewidth,
    width=1.0\linewidth,
    xtick={0,0.2,0.4,0.6,0.8,1.0, 1.2, 1.4},
    legend pos=south west,
    ymajorgrids=true,
    grid style=dashed,
]
\addplot[dashed]coordinates{(0,78.728)(1.4,78.728)};\addlegendentry{SGD(55)}
\addplot[red, mark=triangle] table [x index=0, y index=1, col sep=comma]{data/ptb_large_plot.csv};
\addlegendentry{S-SGD(55)}
\addplot[blue, mark=square] table [x index=0, y index=2, col sep=comma]{data/ptb_large_plot.csv};
\addlegendentry{SGD(29)+S-SGD(26)}
\addplot[orange, mark=o] table [x index=0, y index=3, col sep=comma]{data/ptb_large_plot.csv};
\addlegendentry{SGD(46)+S-SGD(9)}
    \end{axis}
\end{tikzpicture}
\caption{Test perplexities of the large size language model on PTB with different noise scales and training methods. The value inside of the parentheses indicates the training epoch for each method. Please refer to Table 5 in the main paper.}
\label{fig:append_ptb_large}  
\end{figure*}

\begin{figure*}[h]
\centering
\begin{tikzpicture}
\begin{axis}[
    xlabel={Noise level [$\sigma$]},
    ylabel={Test PPL},
    xmin=0, xmax=1.0,
    ymin=82, ymax=92,
    height=0.55\linewidth,
    width=1.0\linewidth,
    xtick={0,0.2,0.4,0.6,0.8,1.0},
    legend pos=south west,
    ymajorgrids=true,
    grid style=dashed,
]
\addplot[dashed]coordinates{(0,88.525)(1.0,88.525)};\addlegendentry{SGD(50)}
\addplot[red, mark=triangle] table [x index=0, y index=1, col sep=comma]{data/ptb_small_plot.csv};
\addlegendentry{S-SGD(50)}
\addplot[blue, mark=square] table [x index=0, y index=2, col sep=comma]{data/ptb_small_plot.csv};
\addlegendentry{SGD(10)+S-SGD(40)}
\addplot[orange, mark=o] table [x index=0, y index=3, col sep=comma]{data/ptb_small_plot.csv};
\addlegendentry{SGD(30)+S-SGD(20)}
    \end{axis}
\end{tikzpicture}
\caption{Test perplexities of small size language model on PTB with different noise scales and training methods. The value inside of the parentheses indicates the training epoch for each method. Please refer to Table 5 in the main paper.}
\label{fig:append_ptb_small}
\end{figure*}

\pagebreak
\section{S-SGD Loss and Derivative}
In the main text, we derive the gradient of S-SGD for a scalar variable case as follows:
\renewcommand\theequation{A.\arabic{equation}}
\begin{align}
\nabla L_{S-SGD} (w_t) &= \frac{\nabla L(w_t+n_t) + \nabla L(w_t-n_t)}{2}\\
             &= \nabla L(w_t) + \frac{\nabla^2 L(w_t + n_{t,1}) - \nabla^2 L(w_t - n_{t,2})}{2} \cdot n
\label{eqn:ap_grad_approximate}
\end{align}
The first term, $\nabla L(w_t)$, is the same as the derivative of SGD loss, while the second term is related to the second order derivatives.  Eq. \ref{eqn:ap_grad_approximate} suggests that if $w_t$ is at the center of a flat minimum in which Hessian terms at $w_{t} + n_{t,1}$ and $w_{t}-n_{t,2}$ are similar, it becomes an optimum point to reach; in contrast, if the Hessian terms at $w_t + n_{t,1}$ and $w_t - n_{t,2}$ are considerably different or have the opposite signs, the weight updates continue even if $\nabla L(w_t)$ is zero.
Here, we extend Eq. \ref{eqn:ap_grad_approximate} for the scalar variable to the multi-dimensional case, where $\mathbf{w},\mathbf{n}  \in \mathbb{R}^D$.  The noise $\mathbf{n}_t$ has a constant strength of $a\sigma(\mathbf{w}_t)$, where $a$ is a predetermined noise level. $\mathbf{n}_{t,1}$ and $\mathbf{n}_{t,2}$ are determined by the Mean-Value theorem.
\begin{align} \centering
\mathbf{\widetilde{w}}_{t,+}=&\mathbf{w}_{t}+\mathbf{n}_{t}, \mathbf{\widetilde{w}}_{t,-}=\mathbf{w}_{t}-\mathbf{n}_{t},\\
\label{eqn:ssgd1}
L(\mathbf{w}_{t}+\mathbf{n}_{t})&=L(\mathbf{w}_{t})+\nabla L(\mathbf{w}_{t}+\mathbf{n}_{t,1})^T \mathbf{n}_{t} \\
L(\mathbf{w}_{t}-\mathbf{n}_{t})&=L(\mathbf{w}_{t})-\nabla L(\mathbf{w}_{t}-\mathbf{n}_{t,2})^T \mathbf{n}_{t} \\
\nabla L_{\text{S-SGD}}(\mathbf{w}_{t}) &=\frac{ \nabla L(\mathbf{w}_{t} + \mathbf{n}_{t})+\nabla L(\mathbf{w}_{t}-\mathbf{n}_{t})}{2} \\
&=\nabla L(\mathbf{w}_{t}) + \frac{1}{2} \left( H_{L} (\mathbf{w}_{t}+\mathbf{n}_{t,1}) - H_{L} (\mathbf{w}_{t}-\mathbf{n}_{t,2}) \right) \mathbf{n}_{t}
\end{align}
Stochastic gradient descent can be considered a random sampling scheme that uses randomly selected mini-batch data. 
Here, we conduct random sampling of two Hessian matrices $H_L(\mathbf{w}_t+\mathbf{n}_{t,1})$ and $H_L(\mathbf{w}_t-\mathbf{n}_{t,2})$ using the noise vector $\mathbf{n}_t$ that has a fixed-magnitude. During the S-SGD training, we generate $\mathbf{n}_t$ to sample the Hessian matrices. Since $\mathbf{n}_t$ has a fixed-magnitude, the result of sampling with noise, $\left( H_{L} (\mathbf{w}_{t}+\mathbf{n}_{t,1}) - H_{L} (\mathbf{w}_{t}-\mathbf{n}_{t,2}) \right) \mathbf{n}_{t}$ is greatly influenced by the largest Hessian terms. Especially, when the Hessian matrices of two separate points, $\mathbf{w}_t+\mathbf{n}_{t,1}$ and $\mathbf{w}_t-\mathbf{n}_{t,2}$ are much different, the noise sampled results can have large values and makes the training keep continue. 
There are several recent researches showing that the largest Hessian values of flat minima have small values when compared with those in sharp minima. When the training guides the network to a flat minimum, the second term has a good possibility of small values. Thus, we can conclude that S-SGD favors flat minima.

\pagebreak
\section{Hessian Eigenvalue Spectrum of SGD and S-SGD Trained Deep Neural Networks}

We plotted the Hessian eigenvalue density in Figure A4, A5 and A6 with the method proposed by \cite{ghorbani2019investigation}. Large eigenvalue suggests that the model is converged to sharp minima. For comparison, we trained WideResNet20x2 on CIFAR-100 with the hyperparameters identical to Section 4. $\lambda$ in the figures denotes the eigenvalue. We can find that the network trained with SGD contains larger Hessian eigenvalues when compared to the one optimized with S-SGD.

\begin{figure}[H]
\centering
\includegraphics[width=0.7\linewidth]{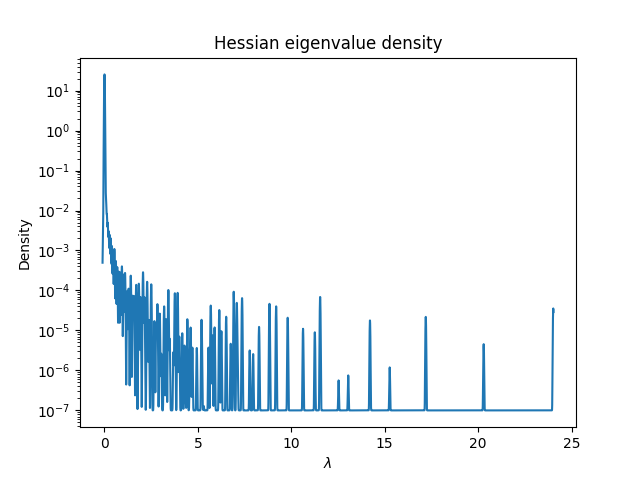} 
\caption{Hessian eigenvalue density of SGD trained WideResNet20x2. The network is trained on CIFAR-100.}
\end{figure}

\begin{figure}[H]
\centering
\includegraphics[width=0.7\linewidth]{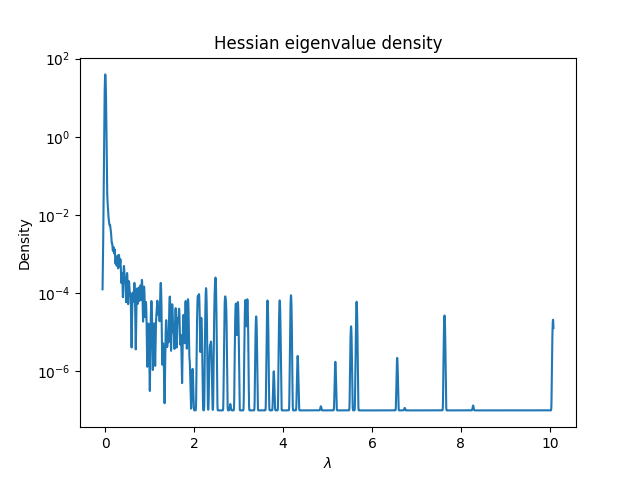} 
\caption{Hessian eigenvalue density of S-SGD trained WideResNet20x2. The network is trained on CIFAR-100.}
\end{figure}
\begin{figure}[H]
\centering
\includegraphics[width=0.95\linewidth]{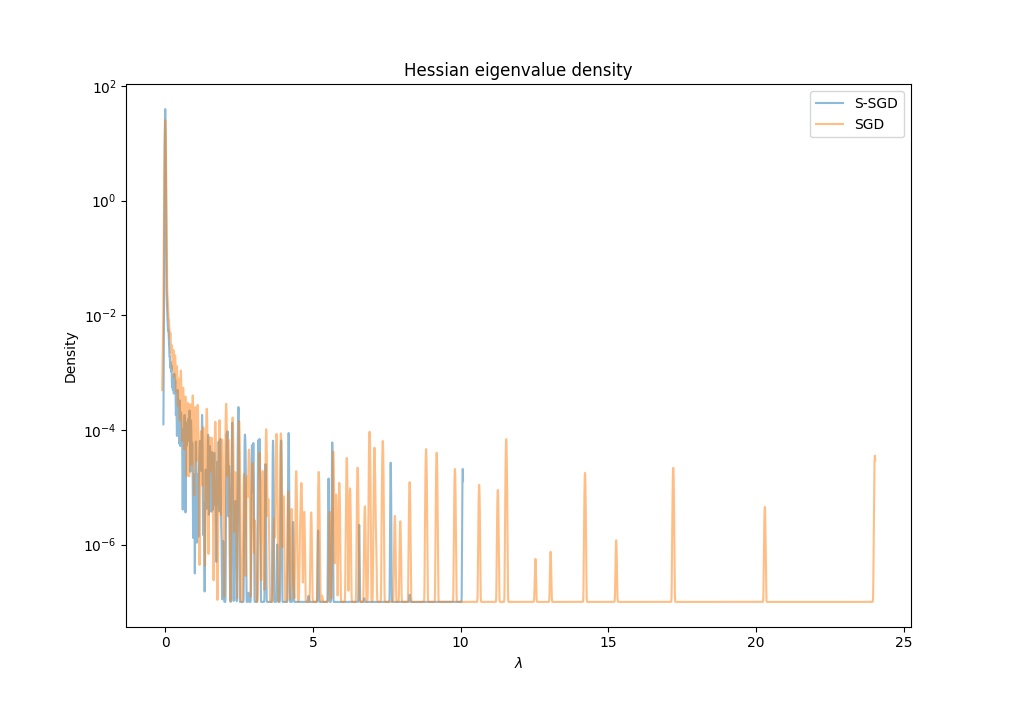} 
\caption{The figure that superimposes Figures A4 and A5.}
\end{figure}
\bibliography{refs}
\bibliographystyle{icml2020}